\documentclass[lettersize,journal]{IEEEtran}
\usepackage{graphicx}
\usepackage{cite}
\usepackage{picinpar}
\usepackage{amsmath}
\usepackage{url}
\usepackage{flushend}
\usepackage[utf8]{inputenc}
\usepackage{colortbl}
\usepackage{soul}
\usepackage{multirow}
\usepackage{pifont}
\usepackage{color}
\usepackage{alltt}
\usepackage[hidelinks]{hyperref}
\usepackage{enumerate}
\usepackage{siunitx}
\usepackage{epstopdf}
\usepackage{pbox}

\usepackage{footmisc}
\usepackage{makecell}

\usepackage[caption=false,font=normalsize,labelfont=sf,textfont=sf]{subfig}

\usepackage{booktabs}

\usepackage{amsmath}   
\usepackage{amssymb}

\usepackage{adjustbox}

\usepackage{amsmath}
\usepackage{booktabs}
\usepackage{graphicx}

\begin{document}
\title{Crisp-Drive: Long-Horizon Consistent and Interaction-Aware World Models for Multi-Style End-to-End Driving}

\author{
	\vskip 0em
		Yuxuan Han$^{1}$, Kunyuan Wu$^{1}$, Liyunong Yang$^{1}$, Zilu Wang$^{1}$, Cansen Jiang$^{2}$, Yi Xiao$^{1}$ and Liang Hu$^{1}$

    	\thanks{$^{1}$ Y. Han, K. Wu, L. Yang, Z. Wang, Y. Xiao and L. Hu are with School of Intelligence Science and Engineering, Harbin Institute of Technology, Shenzhen, China.}
        \thanks{$^{2}$ C. Jiang is with Autonomous Driving Center, Shanghai Utopilot Technology Co.Ltd., China.}
}

\maketitle

% Author names and affiliations are intentionally omitted for T-ASE
% double-anonymous review. Restore them only after acceptance.

\author{
	\vskip 0em
		Yuxuan Han$^{1}$, Kunyuan Wu$^{1}$, Liyunong Yang$^{1}$, Zilu Wang$^{1}$, Cansen Jiang$^{2}$, Yi Xiao$^{1}$ and Liang Hu$^{1}$

    	\thanks{$^{1}$ Y. Han, K. Wu, L. Yang, Z. Wang, Y. Xiao and L. Hu are with School of Intelligence Science and Engineering, Harbin Institute of Technology, Shenzhen, China.}
        \thanks{$^{2}$ C. Jiang is with Autonomous Driving Center, Shanghai Utopilot Technology Co.Ltd., China.}
}

\maketitle

\markboth{IEEE Transactions on Automation Science and Engineering}%
{Anonymous Submission}

\maketitle

\begin{abstract}
World-model-based reinforcement learning has emerged as a promising approach to improving the training efficiency of end-to-end autonomous driving through imagined rollouts. However, existing world models remain limited by temporal inconsistency over long horizons, insufficient modeling of ego-environment interactions, and poor adaptability to diverse driving styles. We propose \textit{Crisp-Drive}, a unified framework that addresses these limitations through temporal consistency modeling, interaction-aware state decomposition, and multi-style policy optimization. First, a gated cross-attention mechanism integrates historical latent states into current predictions, reducing error accumulation and improving the stability of long-horizon imagined rollouts. Second, an explicit state decomposition module separates ego-relevant from ego-irrelevant environmental states, enabling the policy to prioritize safety-critical traffic interactions while suppressing distracting information. Third, Group Relative Policy Optimization employs trajectory-level relative advantages to jointly optimize multiple driving styles and alleviate variations in reward scale. On the Bench2Drive closed-loop benchmark, Crisp-Drive achieves a driving score of 84.12 and a success rate of 63.21\%, outperforming the strongest evaluated world-model baseline by 4.71 points in DS and 10.48 percentage points in SR. Deployment on an automated guided vehicle further demonstrates promising transferability to dynamic real-world driving scenarios.
\end{abstract}

\noindent\textit{Note to Practitioners---}
Automated vehicles must balance safety, efficiency, comfort, and user preferences. Existing learning-based systems often require separate policies for different styles, while long-horizon prediction errors and irrelevant traffic information undermine reliability. Crisp-Drive provides one policy with conservative, moderate, and aggressive modes. It uses recent history and estimates interaction relevance to stabilize decisions and prioritize critical traffic participants. Simulator and low-speed automated guided vehicle tests demonstrate its potential for camera-equipped vehicles while reducing repeated policy training. The current evaluation reports three representative operating points—Conservative, Moderate, and Aggressive—within a continuously parameterized preference space. Future work will investigate online user adaptation and continuous preference adjustment under changing traffic conditions.


\definecolor{limegreen}{rgb}{0.2, 0.8, 0.2}
\definecolor{forestgreen}{rgb}{0.13, 0.55, 0.13}
\definecolor{greenhtml}{rgb}{0.0, 0.5, 0.0}

\section{Introduction}
% \IEEEPARstart{A}{utonomous} driving has increasingly explored end-to-end (E2E) learning paradigms that directly map raw sensor observations to vehicle control actions~\cite{teng2023motion},

Autonomous driving is a critical automation problem in which vehicles operate safely, reliably, and efficiently over long horizons in continuously changing traffic environments. Meeting these requirements demands close coordination between perception and decision-making. End-to-end (E2E) learning therefore offers a promising approach by directly mapping sensor observations to driving actions and enabling system-level joint optimization~\cite{teng2023motion,zhao2025survey,10614862}.
% offering an alternative to traditional modular pipelines comprising perception, prediction, and planning components. 
% This shift is driven by the promise of reducing manual engineering and enabling joint optimization across perception and control, thereby improving adaptability in complex and diverse traffic environments~\cite{zhao2025survey,10614862}.

Existing E2E driving policies are commonly trained through imitation learning (IL) or reinforcement learning (RL)~\cite{hu2023planning,yang2025drivemoe,zhang2021end,yang2025raw2drive}. IL can learn effective behavior from expert demonstrations and typically achieves strong performance under nominal conditions. However, insufficient coverage of rare safety-critical events in demonstrations makes IL policies vulnerable to distribution shift under out-of-distribution (OOD) conditions~\cite{9863660,jaeger2023hidden}. In contrast, RL can improve driving policies through trial-and-error interaction beyond the behaviors contained in demonstrations~\cite{yang2025fast}. Nevertheless, model-free RL generally requires extensive environmental interaction, while direct exploration in driving is costly and potentially unsafe~\cite{gu2025roscom}. This interaction bottleneck motivates world-model-based RL, which learns predictive dynamics from collected experience and uses imagined rollouts to evaluate long-term action consequences and optimize policies with less direct environmental interaction~\cite{guan2024world,pan2023model,hafner2025mastering,zhang2025epona}. By simulating possible futures, the agent can evaluate the long-term consequences of its actions with less direct environmental interaction.

Despite their promise, existing world-model-based driving methods face three coupled challenges. First, autoregressive latent rollouts remain susceptible to error accumulation over long horizons, reducing the reliability of the imagined trajectories used for policy optimization~\cite{zheng2025world4drive}. Second, most models treat surrounding traffic participants uniformly, failing to distinguish ego-critical agents from less relevant ones; this results in suboptimal attention allocation and compromised decision-making~\cite{11080487}. Third, manually weighted step-wise rewards are sensitive to scaling and may inadequately represent trajectory-level driving preferences, increasing the effort required to adapt policies across driving styles~\cite{cai2025multiple}. These challenges are coupled because policy optimization depends on imagined trajectories that are both temporally reliable and focused on decision-critical interactions.

% To tackle these challenges, we propose Crisp-Drive, a novel WM-based RL driving framework designed for robust E2E autonomous driving that supports multiple driving styles. Our key contributions are:

To address these challenges jointly, we propose \textbf{Crisp-Drive}, which combines history-conditioned temporal correction, interaction-relevant state decomposition, and trajectory-level relative policy optimization, enabling style-specific policies to learn from more reliable imagined trajectories that preserve decision-critical interactions.

\begin{enumerate}
     \item We propose a \textbf{Temporal Consistency Regularization} that integrates historical latent context via gated cross-attention to stabilize consistency in long-horizon imagined rollouts and mitigate error compounding.
     \item We introduce an explicit \textbf{State Disentanglement} module that separates ego-relevant from ego-irrelevant interactive states, enabling interpretable and risk-aware decision-making.
     \item We develop a \textbf{Group Relative Policy Optimization} (GRPO) framework that replaces per-step rewards with trajectory-wise relative preferences, and enables joint training of multiple driving styles without reward rescaling.
     \item We perform a practical deployment that demonstrates promising \textbf{sim-to-real transfer} capability of our framework, with robust performance across diverse traffic conditions on the Bench2Drive benchmark and in the real world.
\end{enumerate} 

\section{Related Work}
\subsection{Reinforcement Learning for End-to-End Driving}

Although imitation learning (IL) remains the dominant paradigm for end-to-end driving, its reliance on fixed expert demonstrations makes learned policies
vulnerable to distribution shift during closed-loop deployment. Reinforcement learning (RL) addresses this limitation by optimizing policies through environmental feedback, but direct RL training from raw sensory observations remains unstable and sample-inefficient ~\cite{10614862,wang2026drive}.

Pretrained representations and privileged supervision mainly address the former. For example, GRI~\cite{chekroun2023gri} freezes an encoder learned with auxiliary objectives, while Roach~\cite{zhang2021end} trains a privileged bird's-eye-view (BEV) agent to support sensor-based policy learning, rather than a policy operating directly on raw sensory observations. Meanwhile, safe RL addresses the latter by enforcing constraints during policy training or execution~\cite{alshiekh2018safe}. Runtime mechanisms filter unsafe actions or invoke fallback controllers, whereas constrained optimization incorporates safety conditions through Control Barrier Functions or Lyapunov-based objectives ~\cite{chow2019lyapunov,xie2025cbf}. These mechanisms reduce immediate violations and premature episode termination, but do not explicitly model how complex multi-agent interactions evolve over extended horizons. This limitation motivates world-model-based RL, which augments policy learning with learned predictive dynamics to anticipate long-term consequences before actions are executed ~\cite{10714437,huang2023safedreamer}.
\subsection{World-Model Reinforcement Learning and Interaction-Aware Driving}

World-model RL learns latent environment dynamics and optimizes policies through imagined rollouts~\cite{guan2024world}. RSSM-based methods such as PlaNet~\cite{hafner2019planet} and Dreamer~\cite{hafner2025mastering} establish this framework by predicting future latent states and rewards. In driving, Think2Drive~\cite{li2024think2drive} applies latent imagination to scenario-dense closed-loop training, while Raw2Drive ~\cite{yang2025raw2drive} aligns privileged and raw-sensor world models to guide raw-sensor policy learning, making it directly relevant to raw-sensor MBRL. Adjacent planning approaches include World4Drive~\cite{zheng2025world4drive}, whose intention-aware latent world model generates, evaluates, and selects candidate trajectories, and Epona~\cite{zhang2025epona}, which combines autoregressive future generation and motion planning with spatiotemporal factorization and chain-of-forward training. These methods should be distinguished from standard world-model RL.

% Despite this progress, reliable policy imagination requires both temporal consistency and interaction relevance. Prediction errors can accumulate across long-horizon latent rollouts~\cite{mousakhan2025orbis}, while Iso-Dream~\cite{NEURIPS2022_9316769a} separates controllable and non-controllable dynamics without explicitly identifying which environmental interactions are most relevant to ego decisions. Our method therefore focuses on temporally consistent and ego-relevant latent imagination for policy learning. Given reliable imagined futures, the remaining challenge is to optimize a single policy toward different driving styles.

Despite this progress, reliable policy imagination requires both temporal consistency and interaction relevance. Prediction errors can accumulate across long-horizon latent rollouts~\cite{mousakhan2025orbis}. Iso-Dream~\cite{NEURIPS2022_9316769a} separates controllable and non-controllable dynamics, whereas GraphWorld~\cite{song2026} models ego--agent interactions using an ego-centric graph and improves six-second planning safety. However, jointly maintaining temporal consistency and prioritizing ego-relevant interactions during imagined policy optimization remains unresolved. Our method directly uses temporally regularized and relevance-decomposed latent states to optimize the driving policy through imagined rollouts. Building on this reliable imagination, we further address the challenge of supporting multiple driving styles within a single policy.

\subsection{Style-Conditioned Driving and Relative Policy Optimization}

Most end-to-end driving policies learn a single behavior that reflects the average style of their training demonstrations. Style-conditioned approaches instead expose desired speeds, overtaking preferences, or discrete style variables to the policy. Nevertheless, command conditioning alone does not provide an optimization mechanism for comparing alternative trajectories under different driving styles.

Group Relative Policy Optimization (GRPO)~\cite{shao2024deepseekmath} has recently been adopted for reinforcement post-training in autonomous driving. Poutine~\cite{rowe2025poutine} applies GRPO to a vision-language-trajectory model using a small set of human preference-labeled driving frames. WorldRFT~\cite{yang2026worldrft} performs trajectory-level reinforcement fine-tuning for latent-world-model planning using trajectory Gaussianization and collision-aware GRPO, while EponaV2~\cite{xu2026eponav2} incorporates flow-matching GRPO to improve trajectory planning. These studies demonstrate the effectiveness of relative optimization in driving, but do not jointly optimize multiple behavior styles through interaction-aware latent imagination~\cite{11080487}. Accordingly, our contribution is not the first application of GRPO to autonomous driving; rather, Crisp-Drive couples style-conditioned trajectory-group optimization with an interaction-aware latent world model, enabling a single closed-loop policy to express multiple driving styles without
training separate policies.

\begin{figure*}[!t]
    \centering
    \includegraphics[width=0.9\linewidth]{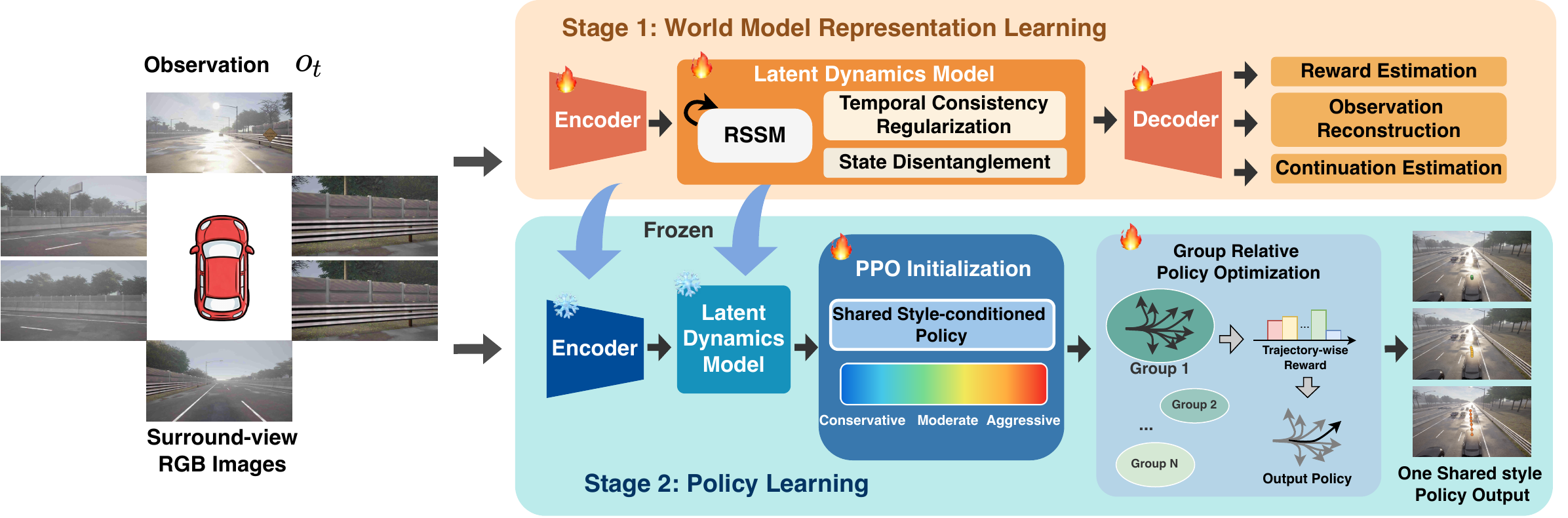}
    \caption{Crisp-Drive follows a two-stage training framework: \textbf{Stage 1) World Model Representation Learning:} an RSSM-based world model encodes observations into a latent state and is trained to reconstruct observations, estimate rewards, and continuation signals; \textbf{Stage 2) Policy Learning:} with the learned world model frozen, shared conditional policy (each representing a distinct driving style) is first pre-trained via PPO and subsequently fine-tuned using GRPO to stabilize learning through intra-group relative advantage estimation.}
    \label{fig_overall}
\end{figure*}

\section{Methodology}
\subsection{Problem Setting}\label{sec:ps}
We formulate WM-based RL for autonomous driving as a partially observable Markov decision process (POMDP). A POMDP is defined by the tuple $(\mathcal{X},\mathcal{A},\mathcal{T},\mathcal{R},\Omega,\mathcal{O},\gamma)$, where $\mathcal{X}$ is the state space of the driving environment; $\mathcal{A}$ is the action space, typically consisting of control signals such as steering angle, throttle, and braking; $\mathcal{T}(x' \,|\, x,a)$ is the state transition function modeling how the environment evolves in response to the agent's actions; the reward function $\mathcal{R}(x, a) \in \mathbb{R}$ provides immediate scalar feedback for taking action $a$ in state $x$; $\Omega$ is the observation space consisting exclusively of multi-view RGB images; $\mathcal{O}(o \,|\, x')$ is the observation function, specifying the probability of perceiving observation $o \in \Omega$ given the underlying state $x' \in \mathcal{X}$; and $\gamma \in [0,1)$ is the discount factor, which balances short-term and long-term rewards. 

At each time step $t$, the agent receives an image observation $o_t\in\Omega$ but does not have direct access to the underlying physical state $x_t$. The observation and action histories define a belief state $b_t=P(x_t\mid o_{1:t},a_{1:t-1})$, which is approximated by a latent state $s_t$ learned by the world model $\mathcal{W}$. Based on this latent representation, the agent selects an action according to a shared style-conditioned policy $a_t\sim\pi_\theta(\cdot\mid s_t,c)$, where $c$ denotes the selected driving-style condition represented by a learned embedding. The style condition remains fixed throughout an episode, while the policy parameters $\theta$ are shared across all styles. Given the condition-dependent reward $r_t(c)$, the policy is optimized to maximize the expected discounted return $J(\theta)=\mathbb{E}[\sum_{t=0}^{T-1}\gamma^t r_t(c)]$.

\subsection{Architecture}\label{sec:archi}
As illustrated in Fig.~\ref{fig_overall}, our framework follows two-stage training procedure comprising World Model Representation Learning and Style-specific Policy Learning. In the first stage, we adopt the Recurrent State Space Model (RSSM)~\cite{hafner2019planet} to construct a world model $\mathcal{W}$ that encodes observations into a latent state $\mathbf{s}_t=[\mathbf{h}_t; \mathbf{z}_t]$, where $\mathbf{h}_t$ is a deterministic recurrent hidden state and $\mathbf{z}_t$ is a stochastic latent variable sampled from a learned distribution. The world model is optimized using observation reconstruction, reward prediction, continuation signals prediction,
% RSSM is trained end-to-end to minimize observation reconstruction error, predict rewards, and continuation signals,
regularized by a Kullback--Leibler (KL) divergence between the prior and posterior distributions over $\mathbf{z}_t$, thereby approximating the POMDP belief state $b_t$. In the second stage, $\mathcal{W}$ is frozen, and multiple driving policies $\pi(a_t \,|\, \mathbf{s}_t)$, each corresponding to a distinct driving style, are first pre-trained using Proximal Policy Optimization (PPO)~\cite{schulman2017proximal} and further fine-tuned via Group Relative Policy Optimization (GRPO)~\cite{shao2024deepseekmath}, which mitigates reward variance through intra-group relative advantage estimation.

\subsubsection{World Model Representation Learning}\label{sec:wml} The world model consists of three core components: an encoder, a latent dynamics model, and a decoder.

\noindent \textbf{Encoder:} At each time step $t$, the agent receives synchronized surround-view images $o_t=\mathbf I_t=\{I_t^{(n)}\}_{n=1}^{N_{\mathrm{cam}}}$, where $I_t^{(n)}\in\mathbb R^{3\times H\times W}$ denotes the $n$-th camera view. A shared VAE encoder $E_{\mathrm{img}}(\cdot)$ extracts a feature map from each view, which is spatially pooled as $\mathbf g_t^{(n)}=\operatorname{GAP}(E_{\mathrm{img}}(I_t^{(n)}))$. To encode the camera viewpoint, its fixed camera-to-ego extrinsic transformation $\mathbf T_n=[\mathbf R_n\mid\mathbf t_n] \in\mathbb R^{3\times4}$ is flattened and projected as $\mathbf e_{\mathrm{ext}}^{(n)}=E_{\mathrm{ext}}(\operatorname{vec}(\mathbf T_n))$, where $E_{\mathrm{ext}}$ is a learnable MLP. The visual and extrinsic features are concatenated along the feature dimension and projected as $\mathbf f_t^{(n)}=\mathbf W_f[\mathbf g_t^{(n)}; \mathbf{e}_{\mathrm{ext}}^{(n)}]$. The camera-aware features are arranged in a fixed camera order and projected into the global observation representation $\mathbf v_t=\mathbf W_v[\mathbf f_t^{(1)}; \ldots; \mathbf f_t^{(N_{\mathrm{cam}})}] \in\mathbb R^{d_v}$.

% During training, an encoder maps the current observation $o_t \in \Omega$ to stochastic representations $\mathbf{z}_t$ for each time step $t$, inferring a posterior distribution over the stochastic latent variable  $\mathbf{z}_t \sim q(\mathbf{z}_t\,|\, o_t, \mathbf{h}_{t})$. In our work, the agent receives a multimodal observation $o_t=(I_t, L_t)$ at each time step $t$, where $I_t$ denotes surround-view camera images and $L_t$ denotes LiDAR point clouds. To obtain a unified representation, we employ a BEVFusion~\cite{10160968} network to project raw sensory inputs $o_t$ into a BEV space, capturing spatial geometry, and a Vision Transformer (ViT)~\cite{DBLP:conf/iclr/DosovitskiyB0WZ21} network to extract high-level visual semantics from $I_t$. The outputs are fused via a linear projection layer $\mathbf{W}$ as $\mathbf{v}_t = \mathbf{W} \left[ \mathbf{BEVFusion}(o_t),\mathbf{ViT}(I_t)\right]$. The posterior distribution over the stochastic latent variable is then inferred as $\mathbf{z}_t \sim q(\mathbf{z}_t\,|\, \mathbf{v}_t, \mathbf{h}_{t})$.

\noindent \textbf{Latent Dynamics Model:} 
We adopt the two-branch RSSM of Iso-Dream++~\cite{pan2023model}, built on the standard RSSM~\cite{hafner2019planet}. It represents the dynamics using a controllable state $\mathbf z_t^{ego}\in\mathbb R^{d_{ego}}$ for action-dependent ego dynamics and an interactive state $\mathbf z_t^{env}\in\mathbb R^{d_{env}}$ for the ego-conditioned global scene context. Their deterministic states are updated by separate Gated Recurrent Unit (GRU):

\begin{equation}
\begin{aligned}
\mathbf{h}^{ego}_t = \text{GRU}_{ego}(\mathbf{h}^{ego}_{t-1}, \mathbf{z}^{ego}_{t-1}, a_{t-1}), \; \\\mathbf{h}^{env}_t = \text{GRU}_{env}(\mathbf{h}^{env}_{t-1}, \mathbf{z}^{ego}_{t-1}, \mathbf{z}^{env}_{t-1}),
\end{aligned}
\end{equation}

The action is supplied directly to the controllable branch, whereas its effect on the interactive dynamics is mediated through the previous controllable state $\mathbf z_{t-1}^{ego}$. The updated deterministic states parameterize the two predictive priors, while the global observation feature $\mathbf v_t$ is used by two branch-specific posterior heads:
\begin{equation}
\begin{aligned}
\hat{\mathbf z}_t^{ego}&\sim_\phi^{ego}(\,\cdot\mid\mathbf h_t^{ego}),&\mathbf z_t^{ego}&\sim_\phi^{ego}(\,\cdot\mid\mathbf v_t,\mathbf h_t^{ego}),\\
\hat{\mathbf z}_t^{env}&\sim_\phi^{env}(\,\cdot\mid\mathbf h_t^{env}),&\mathbf z_t^{env}&\sim_\phi^{env}(\,\cdot\mid\mathbf v_t,\mathbf h_t^{env}).
\end{aligned}
\label{eq:two_branch_prior_posterior}
\end{equation}

All four distributions are diagonal Gaussians parameterized by branch-specific MLPs and sampled using the reparameterization trick.

\begin{figure*}[!t]
    \centering
    \includegraphics[width=0.85\linewidth]{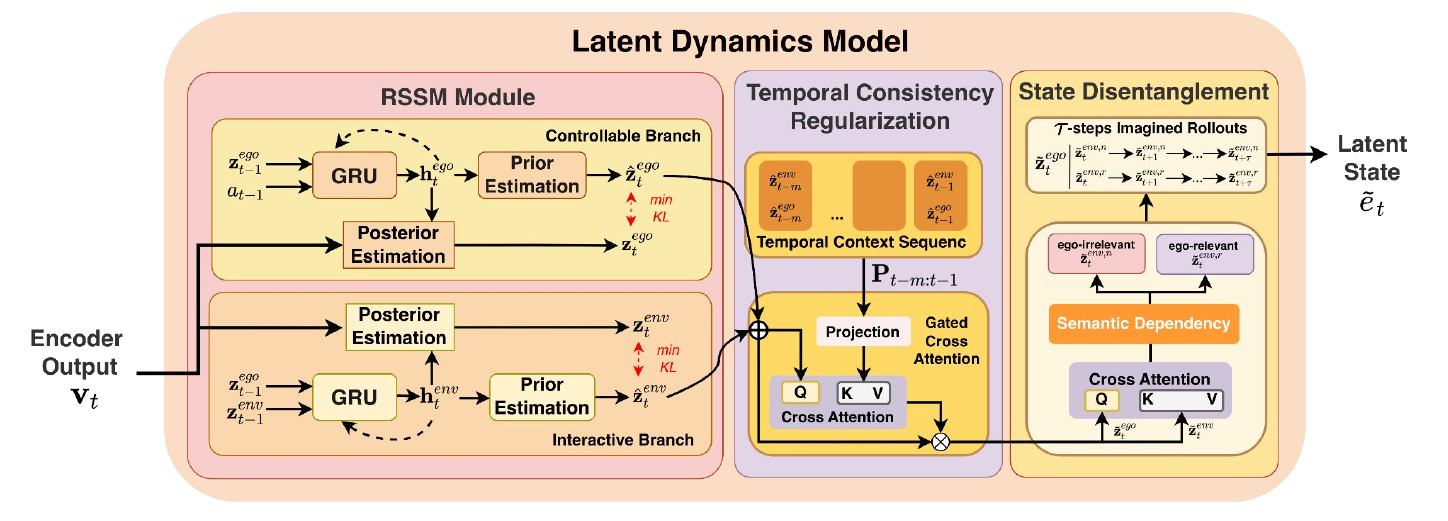}
    \caption{Architecture of the latent dynamics model of our world model: it employs a two-branch RSSM structure to explicitly disentangle controllable (ego) and interactive (environment) states. Temporal consistency regularization integrates historical prior states via gated cross-attention to stabilize long-horizon imagined rollouts. The interactive state is further decomposed into ego-relevant and ego-irrelevant components to facilitate interpretability and safety.}
    \label{fig_representation}
\end{figure*}

A prior sample predicted from only the current recurrent state may be sensitive to local prediction errors during autoregressive imagination. We therefore incorporate the preceding $m$ prior states to refine the current prediction. As illustrated in Fig.~\ref{fig_representation}, the causal history is constructed as
\begin{equation*}
\mathbf P_{t-m:t-1} = \left[ \psi([\hat{\mathbf z}_{t-m}^{ego};
\hat{\mathbf z}_{t-m}^{env}]), \ldots, \psi([\hat{\mathbf z}_{t-1}^{ego}; \hat{\mathbf z}_{t-1}^{env}]) \right],
\end{equation*}
where $\psi(\cdot)$ is a learnable projection. Gated cross-attention~\cite{vaswani2017attention} then integrates the historical prior states with the current prior prediction:
\begin{equation}
\resizebox{0.9\columnwidth}{!}{$\displaystyle 
\begin{bmatrix}\mathbf a_t^{ego}\\ \mathbf a_t^{env}\end{bmatrix}=\operatorname{CrossAtt}\left(Q=\begin{bmatrix}\hat{\mathbf z}_t^{ego}\\ \hat{\mathbf z}_t^{env}\end{bmatrix},K=\mathbf P_{t-m:t-1},V=\mathbf P_{t-m:t-1}\right)$}
\label{eq:history_cross_attention}
\end{equation}

The history-refined prior states are obtained through
\begin{equation}
\begin{aligned}
\begin{bmatrix}
\widetilde{\mathbf z}_t^{ego}\\ \widetilde{\mathbf z}_t^{env} \end{bmatrix} &= \mathbf G_t\odot
\begin{bmatrix} \hat{\mathbf z}_t^{ego}\\ \hat{\mathbf z}_t^{env} \end{bmatrix} + (1-\mathbf G_t)\odot \begin{bmatrix} \mathbf a_t^{ego}\\
\mathbf a_t^{env} \end{bmatrix},\\
\mathbf G_t &= \sigma\left( \phi\left( \begin{bmatrix} [\hat{\mathbf z}_t^{ego};\mathbf a_t^{ego}]\\ [\hat{\mathbf z}_t^{env};\mathbf a_t^{env}] \end{bmatrix} \right) \right),
\end{aligned}
\label{eq:history_refined_states}
\end{equation}
where $\phi(\cdot)$ is an MLP and $\odot$ denotes element-wise multiplication.

To further enhance interpretability and safety, we explicitly disentangle the interactive state $\tilde{\mathbf{z}}^{env}_t$ into \textbf{ego-relevant} and \textbf{ego-irrelevant} components. 
Based on geometric proximity, we first compute spatial attention weights $\mathbf{A}_t = \mathrm{Softmax}\big( \tilde{\mathbf{z}}^{ego}_t \mathbf{W}_Q (\tilde{\mathbf{z}}^{env}_t \mathbf{W}_K)^\top / \sqrt{d} \big)$. The weighted interactive state is then obtained by $\mathbf{\tilde{z}}^{env,loc}_t = \mathbf{A}_t \cdot \tilde{\mathbf{z}}^{env}_t \mathbf{W}_V$, where $\mathbf{W}_Q$, $\mathbf{W}_K$, and $\mathbf{W}_V$ are learnable projections and $d$ is the feature dimension.

Because spatial attention alone may not fully capture semantic interaction relevance, we compute a region-level consistency score:
\begin{equation}
\xi_{t,j}=\frac{1}{2}\left(\frac{\left\langle\widetilde{\mathbf z}^{ego}_t\mathbf W_S^{ego},\widetilde{\mathbf z}^{env}_{t,j}\mathbf W_S^{env}\right\rangle}{\left\|\widetilde{\mathbf z}^{ego}_t\mathbf W_S^{ego}\right\|_2\left\|\widetilde{\mathbf z}^{env}_{t,j}\mathbf W_S^{env}\right\|_2+\epsilon_\xi}+1\right),
\label{eq:semantic_consistency}
\end{equation}
where $\xi_{t,j}\in[0,1]$ measures the semantic consistency between the ego state and the $j$-th spatial region.

We convert this score into a soft relevance probability using,
\begin{equation}
\rho_{t,j} = \sigma\!\left( \frac{\xi_{t,j}-\kappa}{\tau_s} \right),
\label{eq:soft_relevance}
\end{equation}
where $\kappa\in[0,1]$ is the relevance boundary, the sigmoid temperature $\tau_s>0$, and $\sigma(\cdot)$ denotes the sigmoid function. A smaller $\tau_s$ provides a closer approximation to hard thresholding, whereas a larger value produces a smoother relevance transition.

The spatial attention and semantic relevance probability are combined to obtain normalized weights for the relevant and irrelevant regions:
\begin{equation}
\resizebox{0.9\columnwidth}{!}{$
\widehat A^{r}_{t,j}=\frac{A_{t,j}\rho_{t,j}}{\sum_{\ell=1}^{N_{\mathrm{reg}}}A_{t,\ell}\rho_{t,\ell}+\epsilon_A};\widehat A^{n}_{t,j}=\frac{A_{t,j}(1-\rho_{t,j})}{\sum_{\ell=1}^{N_{\mathrm{reg}}}A_{t,\ell}(1-\rho_{t,\ell})+\epsilon_A}.$}
\label{eq:refined_attention}
\end{equation}
where $\widehat A^{r}_{t,j}$ and $\widehat A^{n}_{t,j}$ denote the final aggregation weights assigned to the relevant and irrelevant components, respectively. 

The interactive latent map is then summarized as
\begin{equation}
\resizebox{0.9\columnwidth}{!}{$
\widetilde{\mathbf z}^{env,r}_t=\sum_{j=1}^{N_{\mathrm{reg}}}\widehat A^{r}_{t,j}\left(\widetilde{\mathbf z}^{env}_{t,j}\mathbf W_V\right);
\widetilde{\mathbf z}^{env,n}_t=\sum_{j=1}^{N_{\mathrm{reg}}}\widehat A^{n}_{t,j}\left(\widetilde{\mathbf z}^{env}_{t,j}\mathbf W_V\right),$}
\label{eq:state_decomposition}
\end{equation}
where training-only trajectory annotations are rasterized onto the latent spatial grid to construct region-level relevance targets. 

During each $\tau$-step imagined rollout, the history-refined prior states and actions are generated sequentially. At imagined step $k\in\{0,\ldots,\tau-1\}$, the representation is computed using only the causal prefixes $\widetilde{\mathbf Z}^{env,r}_{t:t+k}=[\widetilde{\mathbf z}^{env,r}_{t},\ldots,\widetilde{\mathbf z}^{env,r}_{t+k}]$ and $\widetilde{\mathbf Z}^{env,n}_{t:t+k}$:
\begin{equation}
\begin{aligned}
\widetilde{\mathbf e}_{t+k}={}&\operatorname{Softmax}\!\left(\widetilde{\mathbf z}^{ego}_{t+k}(\widetilde{\mathbf Z}^{env,r}_{t:t+k})^\top\right)\widetilde{\mathbf Z}^{env,r}_{t:t+k}\\
&+\operatorname{logNorm}\!\left(\widetilde{\mathbf z}^{ego}_{t+k}(\widetilde{\mathbf Z}^{env,n}_{t:t+k})^\top\right)\widetilde{\mathbf Z}^{env,n}_{t:t+k}+\widetilde{\mathbf z}^{ego}_{t+k}.
\end{aligned}
\label{et_method}
\end{equation}
The action $a_{t+k}$ is sampled from $\widetilde{\mathbf e}_{t+k}$ before the next prior state is generated; hence, no action depends on future observations or future posterior states.

\noindent \textbf{Decoder:} Following Iso-Dream++~\cite{pan2023model}, we predict rewards directly from imagined latent states. To support multi-style policy learning, the reward predictor is additionally conditioned on the style representation: $\hat r_t^{\boldsymbol{\alpha}}=\operatorname{MLP}_r([\mathbf e_t;\mathbf c_\alpha])$ and $\hat\gamma_t=\sigma(\operatorname{MLP}_\gamma(\mathbf e_t))$, where $\mathbf c_\alpha$ denotes the encoded style condition introduced in Sec.~\ref{sec:pl}. One VAE decoder is shared across all cameras: 
\begin{equation*} \hat I_t^{(n)}=D_{\mathrm{img}}\!\left(W_d[\mathbf e_t;\mathbf e_{\mathrm{ext}}^{(n)}]\right),\quad n=1,\ldots,N_{\mathrm{cam}}. \end{equation*}

The world model is jointly trained using continuation prediction, reward prediction, observation reconstruction, branch-wise latent regularization, and interaction-relevance supervision. The complete training objective is
\begin{equation*}
\resizebox{0.8\columnwidth}{!}{$\displaystyle
\begin{aligned}
\mathcal L_w &= \mathbb E\sum_{t=1}^{T}\Bigg[\underbrace{-\log p\left(\gamma_t\mid\widetilde{\mathbf z}_t^{ego},\widetilde{\mathbf z}_t^{env}\right)}_{\text{continuation loss}}+\underbrace{\alpha_r\left\|\hat r_t-r_t\right\|_2^2}_{\text{ reward loss}}\\[-0.2em]
&\quad+\underbrace{\alpha_o\left\|\hat{\mathbf I}_t-\mathbf I_t\right\|_2^2}_{\text{ observation reconstruction loss}}\\[-0.2em]
&\quad+\underbrace{\beta_1D_{\mathrm{KL}}\left[q_\phi^{ego}\left(\mathbf z_t^{ego}\mid\mathbf v_t,\mathbf h_t^{ego}\right)\middle\|p_\phi^{ego}\left(\hat{\mathbf z}_t^{ego}\mid\mathbf h_t^{ego}\right)\right]}_{\text{KL: controllable state}}\\[-0.2em]
&\quad+\underbrace{\beta_2D_{\mathrm{KL}}\left[q_\phi^{env}\left(\mathbf z_t^{env}\mid\mathbf v_t,\mathbf h_t^{env}\right)\middle\|p_\phi^{env}\left(\hat{\mathbf z}_t^{env}\mid\mathbf h_t^{env}\right)\right]}_{\text{KL: interactive state}}\Bigg]\\[-0.2em]
&\quad-\underbrace{\frac{\beta_{\mathrm{rel}}}{TN_{ag}}\sum_{s=1}^{T}\sum_{j=1}^{N_{ag}}\left[y_{s,j}\log\rho_{s,j}+(1-y_{s,j})\log(1-\rho_{s,j})\right]}_{\text{interaction-relevance loss}}.
\end{aligned}
$}
\end{equation*}
where $\gamma_t\in\{0,1\}$ is the ground-truth continuation signal, $y_{t,j}\in\{0,1\}$ is the offline relevance pseudo-label of agent $j$, and $\rho_{t,j}$ is its predicted soft relevance probability. The coefficients $\alpha_r$, $\alpha_o$, $\beta_1$, $\beta_2$, and $\beta_{\mathrm{rel}}$ balance the corresponding loss terms.

\subsubsection{Policy Learning}\label{sec:pl}
% In a standard world model framework, a single, fixed-style driving policy is learned by rolling out imagined trajectories in the latent space of the world model. In our work, we aim to learn a \textbf{multi-style driving policy} that can adapt its behavior, such as conservative, moderate, or aggressive, to different driving scenarios or user preferences. To achieve this, we propose a two-stage policy learning approach, as shown in Fig.~\ref{fig:policy}. In the first stage, we train a set of driving policies using imagined rollouts under varying reward coefficients; in the second stage, we further optimize these policies via trajectory-wise relative rewards in parallel interactive environments.

% With the world model frozen, we employ a two-stage style-conditioned policy learning framework. As illustrated in Fig.~\ref{fig:policy}, a reward configuration sampled from a continuous parameter space determines the trajectory reward while its normalized and encoded representation conditions the shared actor and value networks for each rollout. The policy is optimized using PPO and then independently fine-tuned using GRPO with parallel closed-loop rollouts in the same CARLA simulator. Different reward configurations induce diverse driving behaviors, with \textit{Conservative}, \textit{Moderate}, and \textit{Aggressive}
% styles selected as representative operating points in the continuous parameter space.

\begin{figure*}
    \centering
\includegraphics[width=0.9\linewidth]{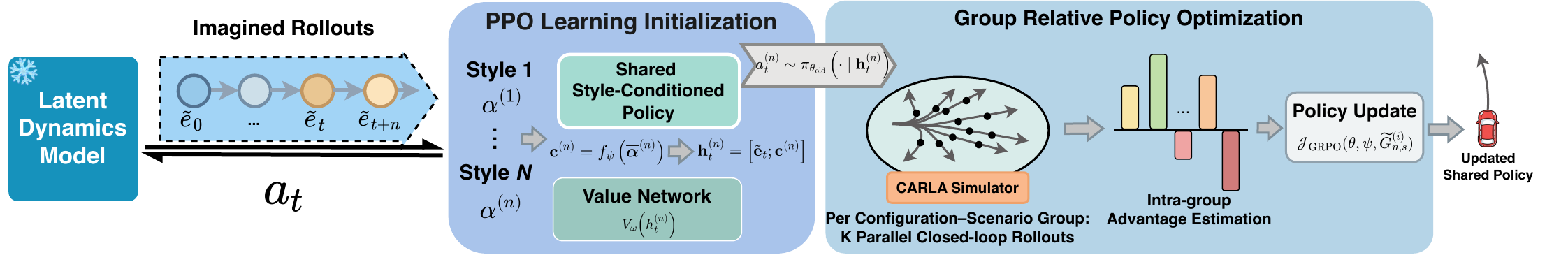}
    \caption{Multi-style policy learning pipeline: A style-conditioned policy is initialized with PPO on imagined rollouts from the frozen world model, with driving styles parameterized by distinct configurations $\boldsymbol{\alpha}^{(n)}$. The policy is then fine-tuned using GRPO, with trajectory-level relative rewards from parallel rollouts.}
    \label{fig:policy}
\end{figure*}

With the world model frozen, we employ a two-stage style-conditioned policy learning framework. As illustrated in Fig.~\ref{fig:policy}, a reward configuration $\boldsymbol{\alpha}$ sampled from a continuous parameter space determines the trajectory reward, while its normalized and encoded representation $\mathbf c_\alpha$ conditions the reward predictor, shared actor, and value networks throughout the same imagined rollout. The policy is optimized using PPO and then independently fine-tuned using GRPO with parallel closed-loop rollouts in the same CARLA simulator. Different reward configurations induce diverse driving behaviors, with \textit{Conservative}, \textit{Moderate}, and \textit{Aggressive} styles selected as representative operating points in the continuous parameter space.

% each policy is first optimized using PPO and then independently fine-tuned using GRPO with parallel closed-loop rollouts in the same CARLA simulator. By assigning a distinct reward configuration to each independently parameterized actor--critic pair, the framework produces \textit{Conservative}, \textit{Moderate}, and \textit{Aggressive} driving styles. All policies share the frozen world model for latent-state inference while maintaining separate actor and critic parameters.

The reward function combines three interpretable objectives:
\begin{equation}
\resizebox{0.9\hsize}{!}{$ 
             R_t = \underbrace{r_{\mathrm{wpt}} + r_{\mathrm{dest}} + r_{\mathrm{time}}}_{\text{Task Completion}}
            + \underbrace{r_{\mathrm{smooth}}}_{\text{Driving Smoothness}}
            + \underbrace{r_{\mathrm{col}} + r_{\mathrm{dev}}}_{\text{Risk Penalties}}.
    $}
\end{equation}
\noindent \textbf{Task Completion:} It comprises a waypoint reward $r_{\mathrm{wpt}}$ that increases linearly with $N_{\mathrm{pass}}$ (\textit{i.e.}, the number of intermediate passed route checkpoints), a destination reward $r_{\mathrm{dest}}$ upon reaching the final goal, 
and a time penalty $r_{\mathrm{time}}$ applied as a fixed negative reward at each time step:
\begin{equation}
\resizebox{0.55\hsize}{!}{$ 
            \begin{aligned}
            r_{\mathrm{wpt}} &= 
                \begin{cases}
                    \alpha_{\mathrm{wpt}} \cdot N_{\mathrm{pass}}, &  N_{\mathrm{pass}} > 0 \\
                    0, & \text{otherwise}
                \end{cases} \\
                r_{\mathrm{dest}} &= 
                \begin{cases}
                    \alpha_{\mathrm{dest}}, & \text{destination reached} \\
                    0, & \text{otherwise}
                \end{cases} \\
                r_{\mathrm{time}} &= -\alpha_{\mathrm{time}},
            \end{aligned}
    $}
\end{equation}
where $\alpha_{\mathrm{wpt}}$, $\alpha_{\mathrm{dest}}$, $\alpha_{\mathrm{time}}$ are reward weights to control the common route-completion objective. 

\noindent \textbf{Driving Smoothness:} which encourages the agent to travel smoothly at a target longitudinal velocity $v_{\mathrm{tgt}}$ while penalizing longitudinal and lateral velocity deviations:
\begin{equation}
\resizebox{0.89\hsize}{!}{$ 
        r_{\mathrm{smooth}} = \alpha_{\mathrm{smooth}} \left( v_{\mathrm{tgt}} - \big| v_{\parallel} - v_{\mathrm{tgt}} \big| - 2\max\left(|v_{\perp}|-0.5,0\right) \right),
    $}
\end{equation}    
where $v_{\parallel}$ and $v_{\perp}$ represent the longitudinal and lateral velocities, respectively. $\alpha_{\mathrm{smooth}}$ is a scalar weight of the driving smoothness reward.

\noindent \textbf{Risk Penalties:} which explicitly discourages hazardous behaviors by a collision penalty $r_{\mathrm{col}}$ and a lane deviation penalty $r_{\mathrm{dev}}$:
\begin{equation}
    \begin{aligned}
    r_{\mathrm{col}} &= 
    \begin{cases}
        -\alpha_{\mathrm{col}} \|\mathbf{v}\|, & \text{collision} \\
        0, & \text{otherwise}
    \end{cases} \\
    r_{\mathrm{dev}} &= 
    \begin{cases}
        -\alpha_{\mathrm{dev}} (d_{\perp} - 0.5), & d_{\perp} > 0.5 \\
        0, & \text{otherwise}
    \end{cases}
    \end{aligned}
\end{equation}
where $\|\mathbf{v}\|$ is ego's speed when collision occurred, and $d_\perp$ indicates the offset distance from the lane centerline. $\alpha_{\mathrm{col}}$ and $\alpha_{\mathrm{dev}}$ are the corresponding reward weights.

The target velocity $v_{\mathrm{tgt}}^{*}$, smoothness coefficient $\alpha_{\mathrm{smooth}}^{*}$, and time-penalty coefficient $\alpha_{\mathrm{time}}^{*}$ are shared across all rollouts. Style-dependent behavior is controlled by the continuously sampled reward configuration,
\begin{equation}
\boldsymbol{\alpha}^{(n)}=\left[\alpha_{\mathrm{col}}^{(n)},\alpha_{\mathrm{dev}}^{(n)},\alpha_{\mathrm{dest}}^{(n)},\alpha_{\mathrm{wpt}}^{(n)}\right],
\label{eq:style_configuration}
\end{equation}
where $s^{(n)}$ is a scalar style variable and $g_{\mathrm{style}}$ maps it to a coupled reward configuration in the calibrated continuous parameter space $\mathcal{A}$. Each sampled configuration remains fixed throughout its rollout, allowing the shared policy to learn different safety--efficiency trade-offs over $\mathcal{A}$. The \textit{Conservative}, \textit{Moderate}, and \textit{Aggressive} styles correspond to three representative evaluation points in this continuous space rather than separately trained policies.

Because the components of $\boldsymbol{\alpha}$ have different units and numerical scales, we normalize them component-wise before policy conditioning. Specifically, $v_{\mathrm{tgt}}$ is linearly normalized using its training bounds, whereas the positive reward coefficients spanning multiple orders of magnitude are log-transformed and min--max normalized:
\begin{equation}
\resizebox{0.94\columnwidth}{!}{$\displaystyle
\overline{\alpha}_j^{(n)}
=
\begin{cases}
\dfrac{\alpha_j^{(n)}-\alpha_j^{\min}} {\alpha_j^{\max}-\alpha_j^{\min}}, & j=\mathrm{tgt},\\[1.0ex]
\dfrac{\log(\alpha_j^{(n)}+\epsilon_{\alpha})-\log(\alpha_j^{\min}+\epsilon_{\alpha})}{\log(\alpha_j^{\max}+\epsilon_{\alpha})-\log(\alpha_j^{\min}+\epsilon_{\alpha})},& j\in\mathcal J_{\mathrm{rw}},\end{cases}\quad\mathcal J_{\mathrm{rw}}=\{\mathrm{col},\mathrm{dev}\}.$}
\label{eq:style_normalization}
\end{equation}

A two-layer style encoder $f_{\psi}$ maps the normalized configuration to a compact embedding:
\begin{equation*}
\mathbf c^{(n)}=f_{\psi}\left(\overline{\boldsymbol{\alpha}}^{(n)}\right)=W_2\sigma\left(W_1\overline{\boldsymbol{\alpha}}^{(n)}+b_1\right)+b_2,
\end{equation*}
where $\sigma(\cdot)$ denotes the SiLU activation. The decision representation $\widetilde{\mathbf e}_t$ produced by the frozen world model is concatenated with the style embedding to form the policy feature,
\begin{equation}
\mathbf h_t^{(n)} = \left[ \widetilde{\mathbf e}_t; \mathbf c^{(n)} \right].
\label{eq:conditioned_policy_feature}
\end{equation}

The shared actor and critic are conditioned on $\mathbf h_t^{(n)}$ as $\pi_{\theta}(a_t\mid\mathbf h_t^{(n)})$ and $V_{\omega}(\mathbf h_t^{(n)})$, respectively. Thus, the sampled raw configuration $\boldsymbol{\alpha}^{(n)}$ parameterizes the trajectory reward, whereas its encoded representation $\mathbf c^{(n)}$ conditions the action distribution and value estimate.

In the first stage, PPO jointly optimizes the shared actor, critic, and style encoder using imagined rollouts generated by the frozen world model. A configuration $\boldsymbol{\alpha}^{(n)}$ is sampled at the beginning of each imagined episode and held fixed throughout the rollout:
\begin{equation}
a_t^{(n)}\sim\pi_{\theta_{\mathrm{old}}}\left(\cdot\mid\mathbf h_t^{(n)}\right),\qquad r_t^{(n)}=R\left(s_t^{(n)},a_t^{(n)};\boldsymbol{\alpha}^{(n)}\right).
\label{eq:conditioned_rollout}
\end{equation}
The conditional critic is trained using generalized advantage estimation, and gradients from all sampled configurations jointly update the shared parameters $(\theta,\omega,\psi)$. Consequently, PPO learns a continuous mapping from the latent driving state and reward preference to the corresponding action distribution and state value.

After PPO initialization, the actor and style encoder are refined using GRPO on parallel closed-loop rollouts. For each sampled configuration $\boldsymbol{\alpha}^{(n)}$ and matched scenario $s$, we collect a group of $K$ trajectories:
\begin{equation}
\displaystyle \mathcal G_{n,s}=\left\{\mathcal T_{n,s}^{(i)}\right\}_{i=1}^{K}, \mathcal T_{n,s}^{(i)}=\left\{(o_t^{(i)},a_t^{(i)},r_t^{(i)})\right\}_{t=1}^{T_i}.
\label{eq:grpo_group}
\end{equation}

All trajectories within $\mathcal G_{n,s}$ share the same configuration and scenario initialization but differ due to action sampling and environment stochasticity. Their discounted returns are $G_{n,s}^{(i)}=\sum_{t=1}^{T_i}\gamma^{t-1}r_t^{(i)}$.

To prevent comparisons across different reward scales and behavioral preferences, the returns are normalized independently within each configuration--scenario group:
\begin{equation}
\resizebox{0.98\columnwidth}{!}{$\displaystyle \mu_{G,n,s}=\frac{1}{K}\sum_{i=1}^{K}G_{n,s}^{(i)},\qquad \sigma_{G,n,s}=\sqrt{\frac{1}{K}\sum_{i=1}^{K}\left(G_{n,s}^{(i)}-\mu_{G,n,s}\right)^2},\qquad \widetilde G_{n,s}^{(i)}=\frac{G_{n,s}^{(i)}-\mu_{G,n,s}}{\sigma_{G,n,s}+\epsilon_G}.$}
\label{eq:group_return_normalization}
\end{equation}
The normalized return is assigned to every step of its trajectory as $\widehat{\mathcal A}_{n,s,i,t}=\widetilde G_{n,s}^{(i)}$. The resulting GRPO objective is
\begin{equation}
\resizebox{0.98\columnwidth}{!}{$\displaystyle
\begin{aligned}
\mathcal J_{\mathrm{GRPO}}(\theta,\psi)&=\mathbb E_{n,s,i,t}\Bigg[\min\left(r_{n,s,i,t}^{\pi}\widehat{\mathcal A}_{n,s,i,t},\operatorname{clip}\left(r_{n,s,i,t}^{\pi},1-\epsilon,1+\epsilon\right)\widehat{\mathcal A}_{n,s,i,t}\right)\\[-0.2em]
&\qquad-\beta_{\mathrm{KL}}D_{\mathrm{KL}}\left[\pi_{\theta}\left(\cdot\mid\mathbf h_t^{(n,i)}\right)\middle\|\pi_{\mathrm{ref}}\left(\cdot\mid\mathbf h_t^{(n,i)}\right)\right]\Bigg].
\end{aligned}
$}
\label{eq:conditional_grpo}
\end{equation}
where $r_{n,s,i,t}^{\pi}=\pi_{\theta}(a_t^{(i)}\mid\mathbf h_t^{(n,i)})/\pi_{\mathrm{old}}(a_t^{(i)}\mid\mathbf h_t^{(n,i)})$ is the conditional importance-sampling ratio. $\pi_{\mathrm{old}}$ denotes the behavior-policy snapshot and is refreshed at the beginning of each GRPO update. In contrast, $\pi_{\mathrm{ref}}$ is a frozen copy of the PPO-initialized policy. Unlike PPO, GRPO uses trajectory-wise relative advantages without critic-based per-step advantage estimation. The losses from all groups jointly update the shared actor and style encoder, while the PPO-trained critic remains fixed during this stage.

\section{Experiments}
\subsection{Experimental Setup}

\subsubsection{Datasets}
We train and evaluate on the Bench2Drive Base subset~\cite{jia2024bench2drive}, built on CARLA~\cite{dosovitskiy2017carla}, and use its official $220$-route protocol for closed-loop testing. We additionally use a self-collected real-world video set only for open-loop rollout evaluation. Dataset scale, collection conditions, splits, and the precise role of each source are documented in Supplementary Material, Sec.~I-A.

\subsubsection{Metrics}
We report the standard Bench2Drive metrics---Success Rate (\textit{SR}), Driving Score (\textit{DS}), Driving Efficiency (\textit{DE}), Driving Comfort (\textit{Comf.}), and five Multi-Ability scores~\cite{jia2024bench2drive}. We additionally evaluate cross-style trajectory diversity and long-horizon visual and geometric consistency using \textit{FID}, \textit{FVD}, \textit{NTA-IoU}, and \textit{NTL-IoU}. Complete definitions, the Diversity equation, normalization, and implementation details are provided in Supplementary Material, Sec.~I-B.

\subsubsection{Training Details}
% The world model is trained for $30$ epochs on $8$ NVIDIA A6000 GPUs with a batch size of $64$, and policy learning uses $100{,}000$ episodes from $K=6$ parallel CARLA environments. Results with uncertainty are reported as mean $\pm$ standard deviation over independently trained runs using the same seed set and evaluation routes for all matched variants. Optimizer schedules, rollout settings, loss weights, seed-control protocol, and all remaining hyperparameters are provided in Supplementary Material, Sec.~I-C.

Bench2Drive is used for world-model training and closed-loop evaluation, with the learned world model providing imagined rollouts for PPO, while the Argoverse training split supplies training-only supervision for interaction-relevance modeling. The world model is trained for $30$ epochs on $8$ NVIDIA A6000 GPUs with a batch size of $16$. The shared style-conditioned policy outputs low-level control commands and is first trained using PPO for $100{,}000$ imagined episodes with an imagination horizon of $\tau=6$, corresponding to $3$ seconds at an effective rollout frequency of $2$~Hz. GRPO then refines the PPO-initialized policy using $K=6$ parallel closed-loop CARLA rollouts for each matched style--scenario group. Crisp-Drive, its ablations, and the reproduced baselines marked by $\dagger$ are evaluated on the same routes using five independent seeds, with results reported as mean $\pm$ standard deviation. Additional implementation details are provided in Supplementary Material, Sec.~I-C.

\subsubsection{Real-world Deployment}
\begin{figure}[!t]
    \centering
    \includegraphics[width=0.9\linewidth]{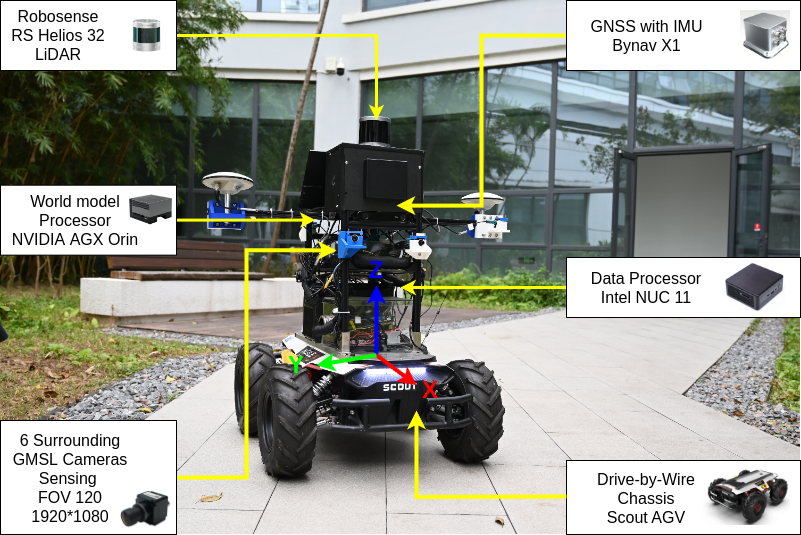}
    \caption{AGV platform for real-world deployment.}
    \label{figure_AGV}
\end{figure}

% We deploy our driving system on a Scout automated guided vehicle (AGV) platform (Fig.~\ref{figure_AGV}) for a closed-loop driving test. The platform is equipped with an NVIDIA Jetson AGX Orin ($64$GB SoC) for real-time policy inference and an Intel NUC 11 for data processing. Sensor suite includes a Bynav X1 GNSS/IMU for localization, a Robosense Helios LiDAR for 3D sensing, and $6$ Sensing SG2-GMSL cameras (each with $1920 \times 1080$ resolution and $120^{\circ}$ HFOV) to provide full-view observations.
We deploy the policy on a Scout automated guided vehicle (AGV), shown in Fig.~\ref{figure_AGV}. An NVIDIA Jetson AGX Orin performs policy inference and data processing. The platform uses a Bynav X1 GNSS/IMU, a RoboSense Helios LiDAR, and six $1920\times1080$ SG2-GMSL cameras with $120^{\circ}$ horizontal fields of view.

\begin{table*}[!t]
\centering
\small
\caption{Performance Comparison of Different E2E Driving Models on Bench2Drive~\cite{jia2024bench2drive}. Results are reported as mean $\pm$ standard deviation over multiple random seeds. Driving conditions (Cond.): IL for imitation learning and RL for reinforcement learning. Input modality: C/L refers to camera/LiDAR. The $\uparrow$ stands for the higher the better. In each metric, the best result among models is shown \textbf{in bold}. $^{\dagger}$ Results reproduced by us using the official
implementation under the same Bench2Drive protocol.}
\label{tab:world_model_comparison}

\setlength{\tabcolsep}{3.5pt} \begin{tabular}{c cc | cccc} \toprule Method & Cond. & Modality & \textit{DS} $\uparrow$ & \textit{SR}(\%) $\uparrow$ & \textit{DE} $\uparrow$ & \textit{Comf.} $\uparrow$ \\ \midrule

UniAD-Base~\cite{hu2023planning} & IL & C & $45.81 \pm {0.51}$ & $16.36 \pm {1.62}$ & $129.21 \pm {2.38}$ & $43.58 \pm {9.15}$ \\

Transfuser~\cite{9863660} & IL & C\&L & $63.58 \pm {3.89}$ & $36.52 \pm {2.96}$ & $88.78 \pm {6.32}$ & $22.96 \pm {2.11}$ \\

Transfuser++~\cite{jaeger2023hidden} & IL & C\&L & $80.87 \pm {0.95}$ & $62.58 \pm {1.02}$ & $125.19 \pm {5.21}$ & $28.17 \pm {3.22}$ \\

DriveMOE~\cite{yang2025drivemoe} & IL & C & $74.22 \pm {2.37}$ & $48.64 \pm {4.35}$ & $175.96 \pm {6.25}$ & $15.31 \pm {1.65}$ \\

GraphWorld~\cite{song2026} & IL & C & $76.70 \pm {5.21}$ & $44.22 \pm {3.65}$ & $80.48 \pm {4.63}$ & $29.74 \pm {2.85}$ \\

ORION~\cite{fu2025orion} & IL & C & $77.74 \pm {3.85}$ & $54.62 \pm {2.57}$ & $151.48 \pm {6.86}$ & $17.38 \pm {2.98}$ \\

\midrule

Iso-Dream++~\cite{pan2023model} & RL & C & $44.54 \pm {1.24}$ & $16.71 \pm {1.96}$ & $170.21 \pm {7.63}$ & $48.64 \pm {3.87}$ \\

World4Drive$^{\dagger}$~\cite{zheng2025world4drive} & RL & C & $64.22 \pm {1.56}$ & $35.08 \pm {2.56}$ & $70.27 \pm {4.85}$ & $16.01 \pm {2.31}$ \\

Epona$^{\dagger}$~\cite{zhang2025epona} & RL & C & $69.23 \pm {2.35}$ & $48.25 \pm {2.71}$ & $171.26 \pm {1.86}$ & $43.58 \pm {1.92}$ \\

Raw2Drive~\cite{yang2025raw2drive} & RL & C & $71.36 \pm {2.78}$ & $50.24 \pm {2.21}$ & $214.17 \pm {3.75}$ & $22.42 \pm {1.25}$ \\

WorldRFT$^{\dagger}$~\cite{yang2026worldrft} & RL & C & $67.35 \pm {1.98}$ & $48.45 \pm {1.91}$ & $151.26 \pm {2.91}$ & $39.56 \pm {3.54}$ \\

EponaV2$^{\dagger}$~\cite{xu2026eponav2} & RL & C & $79.41 \pm {2.95}$ & $52.73 \pm {2.56}$ & $\mathbf{225.33 \pm {3.42}}$ & $28.31 \pm {2.35}$ \\

\midrule

\textbf{Crisp-Drive (Ours)} & RL & C & $\mathbf{84.12 \pm 1.15}$ & $\mathbf{63.21 \pm 1.57}$ & $192.27 \pm {2.56}$ & $\mathbf{49.51 \pm {1.12}}$ \\

\bottomrule
\end{tabular}
\end{table*}

\begin{table*}[!t]
\centering
\small
\caption{Multi-Ability Results on Bench2Drive~\cite{jia2024bench2drive}. The evaluation protocol and highlighting conventions follow Table~\ref{tab:world_model_comparison}.}
\label{tab:world_model_comparison1}

\resizebox{\textwidth}{!}{%
\begin{tabular}{c c c | c c c c c | c}
\toprule
Method & Cond. & Modality & \textit{Merging} $\uparrow$ & \textit{Overtaking} $\uparrow$ & \textit{Emergency Brake} $\uparrow$ & \textit{Give Way} $\uparrow$ & \textit{Traffic Sign} $\uparrow$ & \textit{Mean} $\uparrow$ \\
\midrule
UniAD-Base~\cite{hu2023planning} & IL & C & $14.10 \pm {1.31}$ & $17.78 \pm {1.65}$ & $21.67 \pm {2.36}$ & $10.00 \pm {0.00}$ & $14.21 \pm {2.45}$ & $15.55 \pm {1.63}$ \\
Transfuser~\cite{9863660} & IL & C\&L & $33.07 \pm {3.56}$ & $20.00 \pm {0.00}$ & $35.00 \pm {1.00}$ & $46.67 \pm {4.68}$ & $67.54 \pm {5.87}$ & $40.46 \pm {2.89}$ \\
Transfuser++~\cite{jaeger2023hidden} & IL & C\&L & $41.37 \pm {2.21}$ & $59.62 \pm 7.12$ & $71.42 \pm {8.42}$ & $50.00 \pm {0.00}$ & $68.23 \pm {6.11}$ & $58.13 \pm {3.02}$ \\
DriveMOE~\cite{yang2025drivemoe} & IL & C & $34.67 \pm {2.84}$ & $40.00 \pm {2.00}$ & $65.45 \pm {4.23}$ & $40.00 \pm {1.00}$ & $59.44 \pm {4.36}$ & $47.91 \pm {3.94}$ \\
ORION~\cite{fu2025orion} & IL & C & $25.00 \pm {2.51}$ & $\mathbf{71.11 \pm {4.36}}$ & $78.33 \pm {5.12}$ & $30.00 \pm {0.00}$ & $69.15 \pm {5.92}$ & $54.72 \pm {3.56}$ \\
GraphWorld~\cite{song2026} & IL & C & $42.85 \pm {2.89}$ & $\mathbf{32.57 \pm {2.63}}$ & $49.09 \pm {3.52}$ & $50.00 \pm {0.00}$ & $67.16 \pm {5.27}$ & $48.33 \pm {3.31}$ \\
\midrule
World4Drive~\cite{zheng2025world4drive} & RL & C & $27.38 \pm {2.35}$ & $18.42 \pm {2.85}$ & $35.82 \pm {4.86}$ & $50.00 \pm {2.00}$ & $54.23 \pm {3.45}$ & $37.17 \pm {3.96}$ \\
Iso-Dream++~\cite{pan2023model} & RL & C & $28.82 \pm {2.65}$ & $26.38 \pm {3.68}$ & $48.76 \pm {4.12}$ & $50.00 \pm {5.00}$ & $56.43 \pm {5.20}$ & $42.08 \pm {3.05}$ \\
Epona$^{\dagger}$~\cite{zhang2025epona} & RL & C & $27.13 \pm {1.53}$ & $9.33 \pm {0.72}$ & $20.00 \pm {2.00}$ & $20.00 \pm {3.00}$ & $15.43 \pm {1.12}$ & $14.73 \pm {1.43}$ \\
Raw2Drive~\cite{yang2025raw2drive} & RL & C & $43.35 \pm {2.45}$ & $51.11 \pm {2.56}$ & $60.00 \pm {2.00}$ & $50.00 \pm {0.00}$ & $62.26 \pm {4.59}$ & $53.34 \pm {3.56}$ \\
WorldRFT$^{\dagger}$~\cite{yang2026worldrft} & RL & C & $39.27 \pm {3.12}$ & $46.22 \pm {4.53}$ & $65.00 \pm {0.00}$ & $40.00 \pm {2.00}$ & $56.23 \pm {7.25}$ & $49.34 \pm {3.54}$ \\
EponaV2$^{\dagger}$~\cite{xu2026eponav2} & RL & C & $44.12 \pm {1.72}$ & $54.53 \pm {3.67}$ & $70.37 \pm {2.76}$ & $55.00 \pm {5.00}$ & $63.12 \pm {2.54}$ & $57.43 \pm {5.23}$ \\

\midrule
\textbf{Crisp-Drive (Ours)} & RL & C & $\mathbf{45.22 \pm 1.52}$ & $65.00 \pm {5.00}$ & $\mathbf{80.00 \pm {2.00}}$ & $\mathbf{65.00 \pm {2.00}}$ & $\mathbf{70.00 \pm {0.00}}$ & $\mathbf{65.04 \pm {2.23}}$ \\
\bottomrule
\end{tabular}%
}
\end{table*}

\subsection{Main Experimental Results}
\subsubsection{Quantitative Results}
Tables~\ref{tab:world_model_comparison} and
\ref{tab:world_model_comparison1} compare Crisp-Drive with representative IL- and RL-based E2E driving methods on Bench2Drive. The IL baselines include Transfuser~\cite{9863660}, Transfuser++~\cite{jaeger2023hidden}, DriveMOE~\cite{yang2025drivemoe}, UniAD-Base~\cite{hu2023planning}, and ORION~\cite{fu2025orion}. The RL baselines include World4Drive~\cite{zheng2025world4drive}, Iso-Dream++~\cite{pan2023model}, Raw2Drive~\cite{yang2025raw2drive}, Epona~\cite{zhang2025epona}, WorldRFT~\cite{yang2026worldrft}, and EponaV2~\cite{xu2026eponav2}. As shown in Table~\ref{tab:world_model_comparison}, Crisp-Drive achieves the highest \textit{DS} ($84.12$), \textit{SR} ($63.21\%$), and \textit{Comf.} ($49.51$) among all evaluated methods, despite using only camera observations. It also obtains a competitive \textit{DE} of $192.27$, trailing only EponaV2 and Raw2Drive. Notably, Crisp-Drive outperforms the multi-modal Transfuser++ in all four metrics, demonstrating that its gains do not rely on additional LiDAR input. Compared with the strongest camera-only WM-based baselines, Crisp-Drive provides a better balance among task completion, safety, efficiency, and driving smoothness.

Table~\ref{tab:world_model_comparison1} reports a fine-grained evaluation over five driving abilities. Crisp-Drive achieves the best camera-only results in \textit{Merging} ($45.22$), \textit{Emergency Brake} ($80.00$), \textit{Give Way} ($65.00$), and \textit{Traffic Sign} ($70.00$), as well as the highest single-modal \textit{Mean} score ($65.04$). It also surpasses Epona across all five abilities. Although ORION obtains a higher \textit{Overtaking} score, Crisp-Drive maintains substantially better driving comfort ($49.51$ vs.\ $17.38$), indicating more balanced performance across diverse driving scenarios.

\subsubsection{Qualitative Results}
\begin{figure*}[!t]
    \centering
    \includegraphics[width=0.75\linewidth]{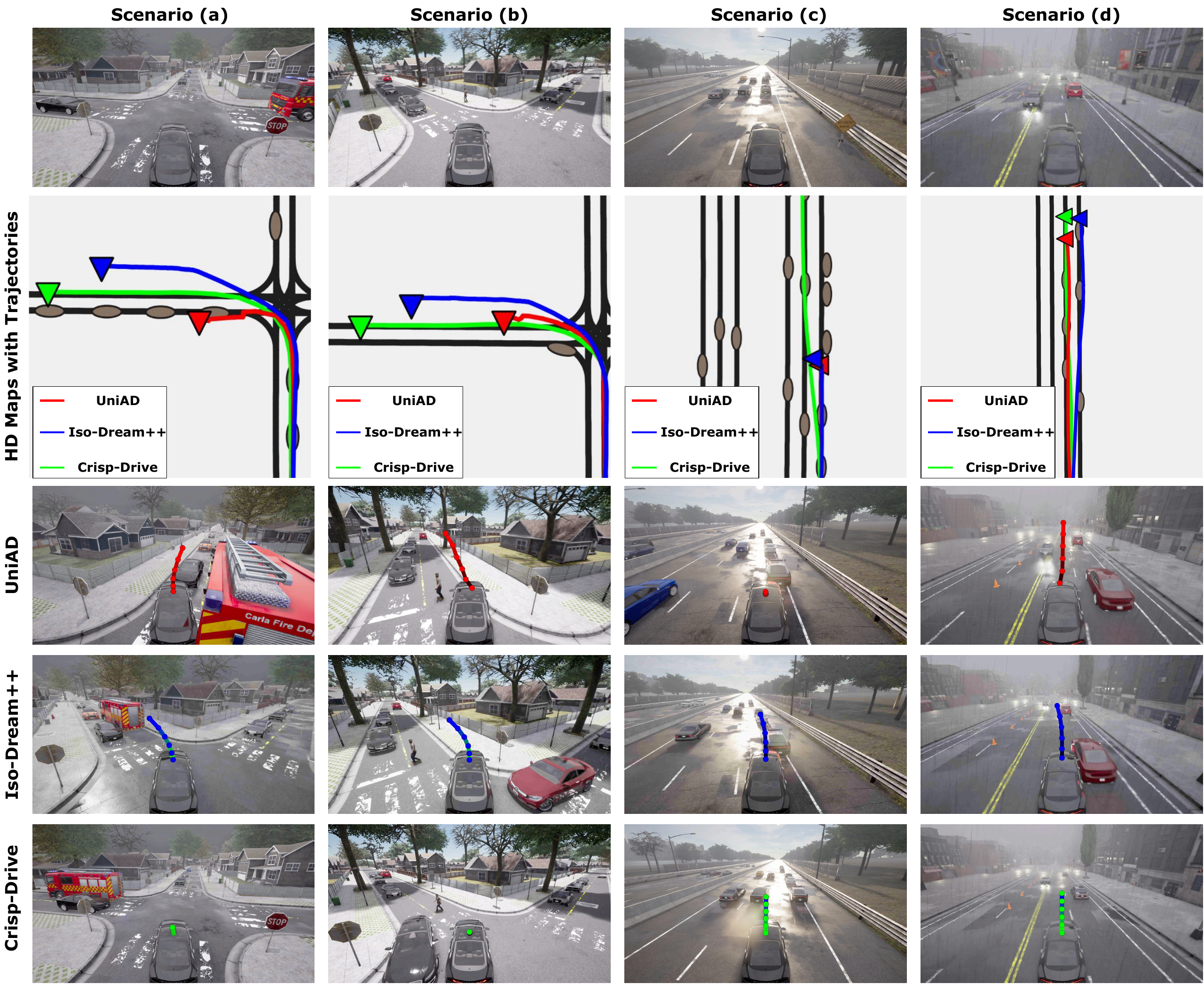}
    \caption{Qualitative comparison of UniAD, Iso-Dream++, and Crisp-Drive in four CARLA scenarios: (a) yielding to an approaching fire truck during merging; (b) yielding to a crossing pedestrian; (c) overtaking unexpectedly stopped traffic; and (d) responding to an oncoming vehicle requesting lane access. Rows show the scene, BEV trajectories, and third-person views; red, blue, and green denote UniAD, Iso-Dream++, and Crisp-Drive, respectively.}
    \label{fig:qualitative_carla}
\end{figure*}

To further analyze the E2E driving performance, we visualize predicted trajectories of UniAD, Iso-Dream++, and Crisp-Drive in four challenging CARLA scenarios (Fig.~\ref{fig:qualitative_carla}): (a) the ego vehicle is about to turn and merge into a target lane while yielding to an approaching fire truck; (b) the ego vehicle is about to turn into a lane where a pedestrian is crossing; (c) the ego vehicle must safely overtake when vehicles ahead stop unexpectedly; (d) the ego vehicle must respond appropriately when an oncoming vehicle from the opposite lane requests to use its lane. Overall, Crisp-Drive demonstrates robust and plausible decision-making in all four scenarios. In Scenario (a), UniAD collides with the fire truck at the intersection, while Iso-Dream++ veers onto the sidewalk aggressively. In contrast, Crisp-Drive correctly yields to the emergency vehicle and safely merges after the truck has passed. Similarly, in Scenario (b), Crisp-Drive yields to the crossing pedestrian while the two baselines fail to stop and exhibit unsafe maneuvers. In Scenario (c), UniAD gets stuck behind the stopping traffic, and Iso-dream++ attempts an unsafe overtake that results in a collision. Crisp-Drive, however, executes a smooth and collision-free lane change to overtake the stopped traffic. In Scenario (d), Crisp-Drive properly negotiates the lane-sharing request, while Iso-Dream++ collides with the vehicle ahead due to its intention to yield to the oncoming vehicle. These results highlight the superior driving performance and behavioral flexibility of our approach.

\subsection{Ablation Study}
% In the ablation study, we evaluate the contribution of three key components in Crisp-Drive: (i) state disentanglement in the latent space, which separates ego-relative and ego-irrelevant dynamics; (ii) the GRPO method, which leverages trajectory-wise relative rewards and enables multi-style policy learning, and (iii) temporal consistency regularization applied during long-horizon imagined rollouts. All ablation experiments are conducted using the multi-modality variant of Crisp-Drive.

We ablate three Crisp-Drive components: temporal consistency regularization, relevance-aware state decomposition, and multi-style GRPO. Unless otherwise noted, all variants use the same camera-only backbone, Bench2Drive routes, random seeds, and matched training budgets. Temporal and relevance variants share the same world-model initialization, while policy-optimization variants start from the same PPO checkpoint.

\begin{table*}[t]
\centering
\caption{Ablation study of the proposed components. The evaluation protocol and highlighting conventions follow Table~\ref{tab:world_model_comparison}}
\label{tab:component_ablation}
\resizebox{\textwidth}{!}{
\begin{tabular}{lccc|cccccc}

\toprule

Setting & Temporal Consistency & Relevance Process & Multi-style GRPO & DS$\uparrow$ & SR$\uparrow$ & DE$\uparrow$ & FID$_{\mathrm{10s}}\downarrow$ & FVD$_{\mathrm{10s}}\downarrow$ & Diversity$\uparrow$ \\

\midrule
Base WM + PPO & $\times$ & $\times$ & $\times$ & $65.23\pm1.32$ & $45.34\pm 1.93$ & $135.34\pm 1.78$ & $90.5$ & $696.8$ & $0.142$ \\

+ Temporal Consistency & $\checkmark$ & $\times$ & $\times$ & $73.10 \pm 0.84$ & $52.46 \pm 1.21$ & $156.82 \pm 3.44$ & $78.20$ & $558.60$ & $0.168$ \\

+ Relevance Process & $\times$ & $\checkmark$ & $\times$ & $75.42 \pm 0.91$ & $54.88 \pm 1.35$ & $168.47 \pm 4.02$ & $86.30$ & $632.50$ & $0.194$ \\

+ Multi-style GRPO & $\times$ & $\times$ & $\checkmark$ & $78.63 \pm 0.73$ & $58.27 \pm 1.12$ & $177.95 \pm 3.67$ & $90.50$ & $696.80$ & $0.452$ \\

\midrule
Full Crisp-Drive & $\checkmark$ & $\checkmark$ & $\checkmark$ & $\mathbf{84.12 \pm 1.15}$ & $\mathbf{63.21 \pm 1.57}$ & $\underline{192.27 \pm {2.56}}$ & $\mathbf{69.4}$ & $\mathbf{428.2}$ & $\mathbf{0.524}$ \\

\bottomrule
\end{tabular}
}
\end{table*}

\noindent\textbf{Overall Component Contributions:}
Table~\ref{tab:component_ablation} isolates the contributions of temporal consistency modeling, relevance-aware state decomposition, and multi-style GRPO. The base WM with PPO achieves a \textit{DS}, \textit{SR}, and \textit{DE} of $65.23$, $45.34\%$, and $135.34$, a 10-second FID/FVD of $90.5$/$696.8$ and a Diversity of $0.142$. Temporal modeling reduces the 10-second FID and FVD by $12.30$ and $138.20$, confirming that historical context mitigates error accumulation during long-horizon imagination. Relevance modeling provides the largest single-component closed-loop gains, improving \textit{DS}, \textit{SR}, and \textit{DE} by $10.19$, $9.54$ percentage points, and $33.13$, respectively, by focusing the policy on decision-critical interactions. Multi-style GRPO primarily improves behavioral diversity from $0.142$ to $0.452$, while also increasing \textit{DS} and \textit{SR} by $13.40$ and $12.93$ percentage points. Combining all components achieves the best overall results, with a \textit{DS} of $84.12$, an \textit{SR} of $63.21\%$, a 10-second FID/FVD of $69.4$/$428.2$, and a Diversity of $0.524$. The consistent improvements over both the baseline and single-component variants demonstrate that the three components provide complementary benefits.

\begin{table}[t]
\centering
\caption{
Ablation study of relevance modeling.
XAttn denotes cross-attention between ego and environment features;
$d$ and TTC denote distance-based and time-to-collision-based relevance
supervision, respectively.
$\uparrow$ indicates that higher is better.
}
\label{tab:relevance_ablation}
\resizebox{\columnwidth}{!}{
\begin{tabular}{c l cccc}
\toprule
& Relevance Setting & $\mathit{DS}\uparrow$ & $\mathit{SR}\uparrow$ & $\mathit{DE}\uparrow$ & $\mathit{Comf.}\uparrow$ \\
\midrule

(a) & None: full $\mathbf{z}_t$ & $54.23$ & $44.25$ & $84.17$ & $16.42$ \\

(b) & $\mathbf{z}_t \rightarrow (\mathbf{z}^{\mathrm{ego}}_t, \mathbf{z}^{\mathrm{env}}_t)$ & $69.12$ & $51.53$ & $119.21$ & $19.01$ \\

(c) & Attn & $73.24$ & $54.17$ & $139.86$ & $27.32$ \\

(d) & $\mathrm{Attn}_{\mathrm{rand}}$ & $67.48$ & $50.36$ & $123.75$ & $21.84$ \\

(e) & $\mathrm{Attn}_{d}$ & $79.16$ & $59.28$ & $175.64$ & $41.73$ \\

(f) & $\mathrm{Attn}_{\mathrm{TTC}}$ & $80.74$ & $60.51$ & $181.92$ & $44.36$ \\

(g) & $\mathrm{Attn}_{d+\mathrm{TTC}}$ \textbf{(Ours)} & $\mathbf{84.12}$ & $\mathbf{63.21}$ & $\mathbf{192.27}$ & $\mathbf{49.51}$ \\
\bottomrule
\end{tabular}
}
\end{table}

\begin{table}[t]
\centering
\caption{Comparison of different temporal modeling architectures under the same history-window length and training budget. $\uparrow$ indicates that higher is better, while $\downarrow$ indicates that lower is better.
}
\label{tab:temporal_architecture}
\resizebox{\columnwidth}{!}{
\begin{tabular}{l ccccc}
\toprule
Variant & $\mathit{DS}\uparrow$ & $\mathit{SR}\uparrow$ & $\mathrm{FID}_{10\mathrm{s}}\downarrow$ & $\mathrm{FVD}_{10\mathrm{s}}\downarrow$ & Lat. (ms)$\downarrow$ \\

\midrule

No history & $70.11$ & $49.23$ & $81.50$ & $549.80$ & $12.40$ \\

Mean pooling & $74.36$ & $53.18$ & $78.62$ & $526.40$ & $12.73$ \\

GRU history & $77.82$ & $56.21$ & $75.48$ & $493.65$ & $13.58$ \\

Cross-attention w/o gate & $81.05$ & $60.02$ & $72.31$ & $458.74$ & $14.26$ \\

Gated cross-attention (Ours) & $\mathbf{84.12}$ & $\mathbf{63.21}$ & $\mathbf{69.40}$ & $\mathbf{428.20}$ & $14.61$ \\

% \midrule
% Gated cross-attention (Ours) & $\mathbf{84.12 \pm 0.15}$ & $\mathbf{63.21 \pm 0.57}$ & \textbf{--} & \textbf{--} & -- \\
\bottomrule
\end{tabular}
}
\end{table}

\begin{table}[t]
\centering
\caption{
Comparison of policy optimization mechanisms.
$\uparrow$ indicates that higher is better.
}
\label{tab:policy_optimization}
\resizebox{\columnwidth}{!}{
\begin{tabular}{l cc cccc}
\toprule
Setting & Group Norm. & KL & $\mathit{DS}\uparrow$ & $\mathit{SR}\uparrow$ & $\mathit{Comf.}\uparrow$ & Diversity$\uparrow$ \\
\midrule
PPO & -- & -- & $65.07$ & $49.12$ & $20.55$ & $0.151$ \\

PPO-longer & -- & -- & $67.84$ & $51.07$ & $24.63$ & $0.173$ \\

Trajectory-return PPO & $\times$ & $\checkmark$ & $73.26$ & $55.42$ & $34.17$ & $0.314$ \\

GRPO w/o KL & $\checkmark$ & $\times$ & $80.48$ & $60.17$ & $45.82$ & $0.468$ \\

\midrule
Full GRPO & $\checkmark$ & $\checkmark$ & $\mathbf{84.12}$ & $\mathbf{63.21}$ & $\mathbf{49.51}$ & \textbf{0.524} \\
\bottomrule
\end{tabular}
}
\end{table}

\noindent \textbf{State Disentanglement:}
Table~\ref{tab:relevance_ablation} evaluates whether the gains arise from meaningful relevance supervision rather than additional capacity. Settings (a)--(b) compare the full latent state with the basic ego--environment split; (c)--(d) use unsupervised attention and random labels; and (e)--(g) examine distance, TTC, and their combination.

The ego--environment split improves \textit{DS} and \textit{SR} to $69.12$ and $51.53\%$. Unsupervised and random-label attention remain inferior to the semantically supervised variants, excluding model capacity and arbitrary partitioning as the primary causes of improvement. Combining distance and TTC performs best, achieving $84.12$ \textit{DS}, $63.21\%$ \textit{SR}, $192.27$ \textit{DE}, and $49.51$ \textit{Comf.}

% \begin{table}[t]
% \centering
% \caption{Ablation study on the impact of different policy optimization strategies on the driving robustness and diversity. The experiments are based on Bench2Drive~\cite{jia2024bench2drive}. The best result is shown \textbf{in bold}.}
% \label{tab_policy_ablation}
% \resizebox{\columnwidth}{!}{
% \begin{tabular}{cccc ccccc}
% \toprule
% &\multicolumn{2}{c}{\textbf{Policy Optimization Strategy}} & \multirow{3}{*}{\textit{DS} $\uparrow$} & \multirow{3}{*}{\textit{SR} $\uparrow$ } & \multirow{3}{*}{\textit{DE} $\uparrow$} & \multirow{3}{*}{\textit{Comf.} $\uparrow$} & \multirow{3}{*}{Diversity $\uparrow$}\\
% \cmidrule(r){2-3}
% & Multiple $\boldsymbol{\alpha}^{(n)}$ & GRPO & & & & &\\
% &  ($N > 1$)  & Fine-tuning & & & & &\\
% \midrule
% (a) & \texttimes & \texttimes & $65.07$ & $49.12$ & $94.91$ & $20.55$ & $0.151$\\
% (b) & \checkmark & \texttimes & $67.12$ & $55.33$ & $112.51$ & $23.18$ & $0.312$\\
% (c) & \texttimes & \checkmark & $74.21$ & $62.12$ & $152.71$ & $35.29$ & $0.193$\\
% (d) & \checkmark & \checkmark &  $\textbf{88.44}$ & $\textbf{66.82}$ & $\textbf{174.42}$ & $\textbf{40.92}$ & \textbf{0.524} \\
% \bottomrule
% \end{tabular}
% }
% \end{table}

To gain insight into the internal representations of the WMs, we visualize the attention maps of Crisp-Drive and Iso-Dream++ in two interactive scenarios (see Fig.~S1 in the Supplementary Material). Crisp-Drive assigns significantly higher attention weights to ego-relevant agents within the interactive state (\textit{e.g.}, crossing pedestrians or oncoming vehicles requesting lane access). In contrast, Iso-Dream++ exhibits less discriminative attention patterns, distributing attention more uniformly across surrounding entities.

% \noindent \textbf{GRPO Method:} 
% In Table~\ref{tab_policy_ablation}, we compare our proposed policy optimization strategy (setting (d)) against three baselines: (a) standard PPO trained with scalar rewards to learn a fixed-style driving policy; (b) PPO trained with scalar rewards to learn multi-style driving policies; (c) PPO fine-tuned via GRPO but constrained to a single driving style. 

% The baseline (a) exhibits the lowest performance across all metrics, particularly in \textit{Diversity} ($0.151$), underscoring the limitations of conventional RL in generating diverse driving behaviors. Introducing multi-style policies, baseline (b) improves \textit{DS} ($+2.05$), \textit{SR} ($+6.21\%$), and achieves a significantly higher score in \textit{Diversity} ($0.312$, twice over (a)), demonstrating that explicit multi-style PPO enables richer driving behaviors while maintaining safety. Applying GRPO fine-tuning ((c) and (d)) leads to substantial gains in task performance. Even with one style, GRPO boots \textit{DS} by $+9.14$ and \textit{SR} by $+13.00\%$ over (a), and notably enhances \textit{DE} ($+57.80$) and \textit{Comf.} ($+14.74$), reflecting the benefit of trajectory-wise relative rewards. Our model integrating multi-style policies and GRPO fine-tuning delivers the best results across all metrics, with \textit{Diversity} reaching $0.524$ ($+68\%$ over (b)).

\noindent\textbf{Policy Optimization:}
Table~\ref{tab:policy_optimization} compares policy-optimization mechanisms under the same PPO initialization and additional interaction budget. PPO-longer controls for extra training, Trajectory-return PPO isolates the effect of trajectory-level returns, and GRPO without KL evaluates the contribution of policy regularization. Standard PPO obtains a \textit{DS} of $65.07$, an \textit{SR} of $49.12\%$, a \textit{Comf.} of $20.55$, and a Diversity of $0.151$. Extending PPO training improves these metrics only to $67.84$, $51.07\%$, $24.63$, and $0.173$, indicating that additional updates alone are insufficient. Trajectory-return PPO further increases \textit{DS} and \textit{SR} by $5.42$ and $4.35$ percentage points, while group normalization provides additional gains of $10.86$ and $7.79$ percentage points in \textit{DS} and \textit{SR}, respectively, by reducing return-scale variation. Removing KL decreases \textit{DS} from $84.12$ to $80.48$ and Diversity from $0.524$ to $0.468$, showing its importance for stable fine-tuning. Full GRPO improves over PPO by $19.05$ in \textit{DS}, $14.09$ percentage points in \textit{SR}, and $28.96$ in \textit{Comf.}, while increasing Diversity by $0.373$.

\begin{table*}[t]
\centering
\caption{Ablation study on the impact of temporal consistency regularization, evaluated in closed-loop driving (Bench2Drive~\cite{jia2024bench2drive}) and open-loop imagined rollouts (Bench2Drive evaluation set and self-collected real-world data) across four prediction horizons at $1$, $3$, $5$, and $10$ seconds. $\uparrow$ indicates that higher is better, while $\downarrow$ indicates that lower is better. The best result is shown in \textbf{bold}.}
\label{tab:long_term_ablation}
\resizebox{\textwidth}{!}{
\begin{tabular}{l cccc | cc cc cc cccc}
\toprule
\multirow{3}{*}{Method} & \multicolumn{4}{c|}{Closed-loop Driving Test} & \multicolumn{10}{c}{Open-loop Imagined Rollouts Test} \\
\cmidrule(lr){2-5} \cmidrule(lr){6-15} & \multicolumn{4}{c|}{} & \multicolumn{2}{c}{1s} & \multicolumn{2}{c}{3s} & \multicolumn{2}{c}{5s} & \multicolumn{4}{c}{10s} \\
\cmidrule(lr){6-7} \cmidrule(lr){8-9} \cmidrule(lr){10-11} \cmidrule(lr){12-15} & \textit{DS}$\uparrow$ & \textit{SR}$\uparrow$ & \textit{DE}$\uparrow$ & \textit{Comf.}$\uparrow$ & \textit{FID}$\downarrow$ & \textit{FVD}$\downarrow$ & \textit{FID}$\downarrow$ & \textit{FVD}$\downarrow$ & \textit{FID}$\downarrow$ & \textit{FVD}$\downarrow$ & \textit{FID}$\downarrow$ & \textit{FVD}$\downarrow$ & \textit{NTA-IoU}$\uparrow$ & \textit{NTL-IoU}$\uparrow$ \\

\midrule
DreamerV3~\cite{hafner2025mastering} & -- & -- & -- & -- & 45.8 & $452.0$ & $72.3$ & $573.2$ & $90.5$ & $696.8$ & $145.1$ & $879.3$ & $0.102$ & $32.214$ \\

Iso-Dream++~\cite{pan2023model} & $44.54 \pm 1.24$ & $16.71 \pm 1.96$ & $170.21 \pm 7.63$ & $48.64 \pm 0.72$ & $29.7$ & $182.3$ & 45.1 & 399.2 & 59.3 & $462.7$ & $91.7$ & $552.3$ & $0.115$ & $36.786$ \\

DriveDreamer~\cite{wang2024drivedreamer} & -- & -- & -- & -- & $40.5$ & $217.9$ & $59.7$ & $427.9$ & $71.4$ & $517.3$ & $99.9$ & $585.8$ & $0.175$ & $40.112$ \\

World4Drive~\cite{zheng2025world4drive} & $64.22 \pm {1.56}$ & $35.08 \pm 2.56$ & $70.27 \pm 4.85$ & $16.01 \pm 2.31$ & $21.3$ & $177.5$ & $39.8$ & $349.7$ & $52.7$ & $452.1$ & $88.3$ & $567.2$ & $0.217$ & $43.323$ \\

Epona~\cite{zhang2025epona} & $69.23 \pm {2.35}$ & $48.25 \pm {2.71}$ & $171.26 \pm {1.86}$ & $43.58 \pm {1.92}$ & $15.8$ & $157.2$ & $35.2$ & $311.6$ & $47.7$ & $422.4$ & $83.4$ & $572.4$ & $0.234$ & $48.331$ \\
\midrule

\textbf{Ours} (\textbf{w/o} Temporal Reg.) & $70.11 \pm 1.23$ & $49.23\pm 1.89$ & $142.13\pm2.03$ & $23.51 \pm 0.77$ & $17.2$ & $122.7$ & $37.6$ & $321.2$ & $67.9$ & $502.1$ & $81.5$ & $549.8$ & 0.231 & 50.113 \\

\textbf{Ours} (\textbf{with} Temporal Reg.) & $\mathbf{84.12 \pm 1.15}$ & $\mathbf{63.21 \pm 1.57}$ & $\mathbf{192.27 \pm 2.56}$ & $\mathbf{49.51 \pm {1.12}}$ & $\mathbf{12.3}$ & $\mathbf{99.3}$ & $\mathbf{24.7}$ & $\mathbf{257.3}$ & $\mathbf{46.5}$ & $\mathbf{372.5}$ & $\mathbf{69.4}$ & $\mathbf{428.2}$ & $\mathbf{0.392}$ & $\mathbf{55.122}$ \\
\bottomrule
\end{tabular}
}
\end{table*}

\begin{table}[t]
\centering
\caption{
Style-conditioned Multi-Ability evaluation on the Bench2Drive
closed-loop benchmark. All driving styles are evaluated on the same
routes and random seeds.
}
\label{tab:style_multi_ability}
\resizebox{\columnwidth}{!}{
\begin{tabular}{c cccccc}
\toprule
Style & Merging$\uparrow$ & Overtaking$\uparrow$ & Emergency Brake$\uparrow$ & Give Way$\uparrow$ & Traffic Sign$\uparrow$ & Mean$\uparrow$ \\
\midrule
$\mathcal{C}$ & $38.12 \pm 2.50$ & $45.00 \pm 5.00$ & $80.00 \pm 2.5$ & $65.00 \pm 5.00$ & $72.00 \pm 1.00$ & $60.02 \pm 2.50$ \\

$\mathcal{M}$ & $45.22 \pm 1.52$ & $65.00 \pm {5.00}$ & $80.00 \pm {2.00}$ & $65.00 \pm {2.00}$ & $70.00 \pm {0.00}$ & $65.04 \pm {2.23}$ \\

$\mathcal{A}$ & $50.80 \pm 3.00$ & $75.00 \pm 5.00$ & $70.00 \pm 3.00$ & $45.00 \pm 5.00$ & $65.00 \pm 4.00$ & $61.16 \pm 2.10$ \\
\bottomrule
\end{tabular}
}
\end{table}

\begin{figure*}
    \centering
    \includegraphics[width=0.75\linewidth]{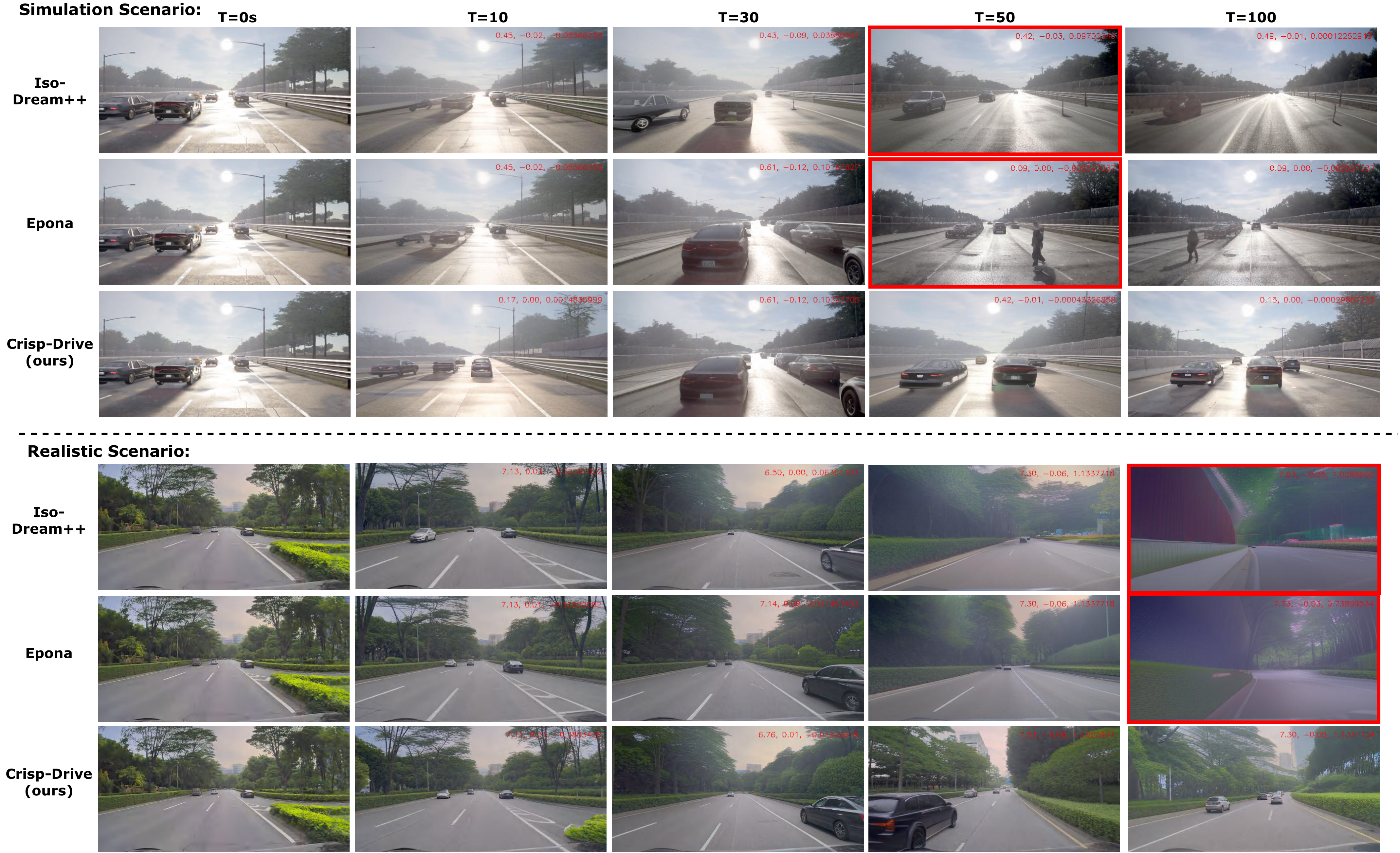}
    \caption{Qualitative comparison of imagined rollouts from Iso-Dream++, Epona, and Crisp-Drive in both the CARLA simulator and real-world driving environments. Frames exhibiting temporally inconsistent predictions are highlighted with red bounding boxes.}
    \label{figure_realworld_worldmodel}
\end{figure*}

\noindent \textbf{Temporal Consistency Regularization:}
Results in Table \ref{tab:long_term_ablation} demonstrate the critical role of temporal consistency regularization in enhancing both closed-loop driving performance and open-loop rollout fidelity. For a comprehensive comparison, we also report the results of five representative WMs
% \footnote{Certain baselines report only results on open-loop imagined rollouts, as they were not originally designed for the E2E closed-loop driving task.}:
DreamerV3~\cite{hafner2025mastering}, Iso-Dream++~\cite{pan2023model}, DriveDreamer~\cite{wang2024drivedreamer}, World4Drive~\cite{zheng2025world4drive}, and Epona~\cite{zhang2025epona}. In closed-loop evaluation on Bench2Drive, Crisp-Drive with temporal regularization achieves substantial improvements over the variant without it, boosting \textit{DS}($+14.01$), \textit{SR} ($+13.98\%$), \textit{DE} ($+50.14$), and \textit{Comf.} ($+26.00$). These results highlight the positive impact of temporal regularization on driving safety, task completion, and driving smoothness. Compared to other WMs, our regularized model achieves the highest \textit{DS} and \textit{SR}, surpassing the existing strongest baseline EponaV2 by $+4.71$ and $+10.48\%$, respectively. In open-loop imagined rollouts across $1$-$10$ seconds, our model consistently attains the lowest \textit{FID} and \textit{FVD} scores among all baselines. Notably, it outperforms Epona by margins that grow with horizon length: we achieve \textit{FID} reductions of $3.5$, $10.5$, $1.2$, and $14.0$, and \textit{FVD} reductions of $57.9$, $54.3$, $49.9$, and $144.2$ at $1$s, $3$s, $5$s, and $10$s, respectively. Crucially, beyond the $5$-second horizon where most methods exhibit accelerating error accumulation (evidenced by rapid \textit{FVD} growth), our model shows significantly slower degradation in both metrics. 

\begin{table}[t]
\centering
\caption{Ablation study of the imagination horizon.}
\label{tab:imagination_horizon}
\resizebox{\columnwidth}{!}{
\begin{tabular}{c c c c c}
\toprule
$\tau$ & \textit{DS}$\uparrow$ & \textit{SR}$\uparrow$ & \textit{DE}$\uparrow$ & Latency (ms)\\
\midrule
3  & $77.12 \pm 0.23$ & $60.17 \pm 0.77$ & $172.11 \pm 2.73$ & $115$ \\
6  & $84.12 \pm 1.15$ & $63.21 \pm 1.57$ & $192.27 \pm 2.56$ & $231$ \\
9  & $\mathbf{88.43 \pm 1.23}$ & $\mathbf{66.33 \pm 1.13}$ & $\mathbf{201.77 \pm 3.32}$ & $473$ \\
12 & $86.23 \pm 0.75$ & $65.57 \pm 1.45$ & $199.34 \pm 3.43$ & $723$ \\
\bottomrule
\end{tabular}
}
\end{table}

To facilitate intuitive comparison, Fig.~\ref{figure_realworld_worldmodel} presents imagined rollouts from Iso-Dream++, Epona, and Crisp-Drive in both CARLA and real-world driving environments, showing key frames at timesteps $0$, $10$, $30$, $50$, and $100$. Iso-Dream++ and Epona frequently produce inconsistent predictions, such as physically implausible transitions, visual artifacts, and color distortions; in contrast, Crisp-Drive maintains temporal coherence over long-horizon rollouts in both domains.

\noindent\textbf{Multi-Style Driving Ability:} Table~\ref{tab:style_multi_ability} evaluates the three driving styles on
identical Bench2Drive routes. The Conservative, Moderate, and Aggressive policies achieve \textit{Mean} scores of $60.02$, $65.04$, and $61.16$, respectively, showing that all styles preserve competitive task competence. The Conservative policy obtains higher \textit{Emergency Brake} and \textit{Give Way} scores of $80.00$ and $65.00$, exceeding the Aggressive policy by $10.00$ and $20.00$, respectively. Conversely, the Aggressive policy improves \textit{Merging} and \textit{Overtaking} by $12.68$ and $30.00$, respectively, over the Conservative policy. The Moderate policy remains between the two extremes across most abilities, demonstrating stable and interpretable style-conditioned behavior.

\begin{figure*}[t]
\centering

\subfloat[Scenario 1: oncoming obstacle avoidance%
\label{fig:sub1}]{
    \includegraphics[width=0.8\textwidth]{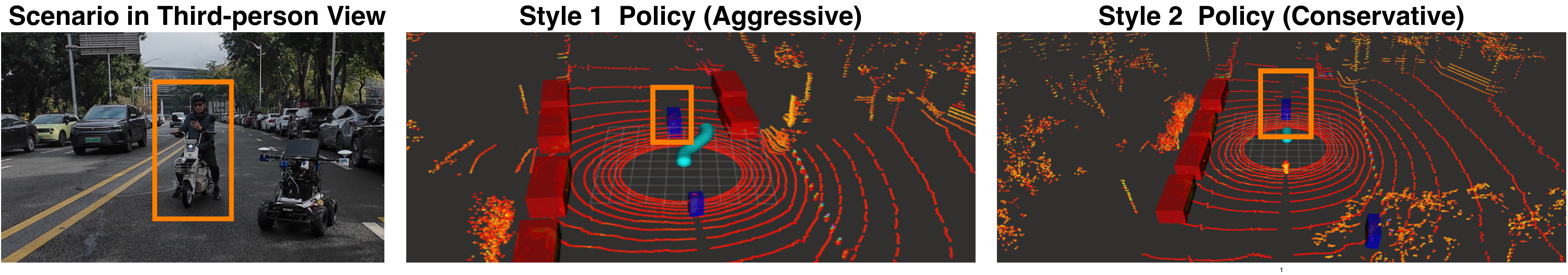}
}
\par
\subfloat[Scenario 2: mid-road static obstacle avoidance%
\label{fig:sub2}]{
    \includegraphics[width=0.8\textwidth]{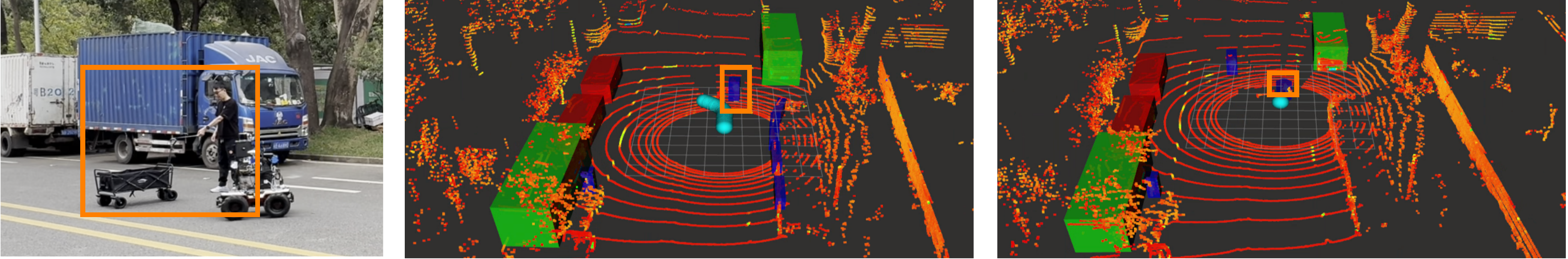}
}
\par

\subfloat[Scenario 3: emergency response to jaywalking pedestrians%
\label{fig:sub3}]{
    \includegraphics[width=0.8\textwidth]{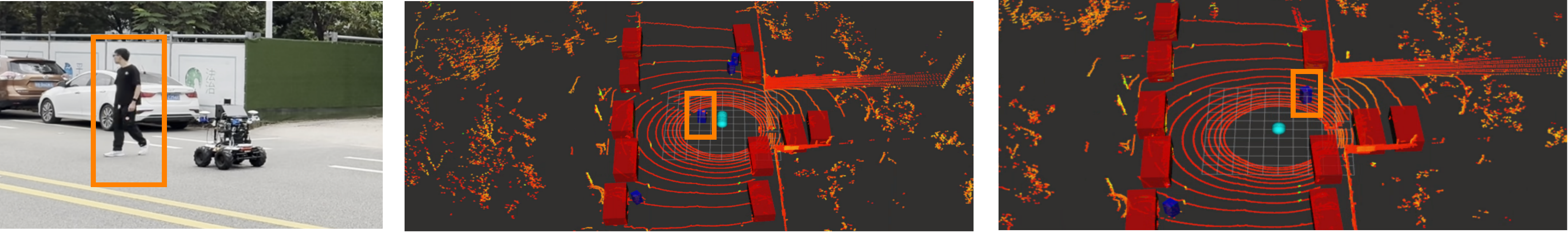}
}

\caption{From top to bottom, we show the E2E driving results of Crisp-Drive deployed on an AGV in three real-world scenarios. In each row, we illustrate a third-person view of the scenario, the output trajectories under two distinct driving styles, and the corresponding LiDAR BEV maps.}
\label{fig:qualitative}

\end{figure*}

\noindent\textbf{Imagination Horizon:}
Table~\ref{tab:imagination_horizon} analyzes the rollout length for imagined rollouts. Increasing $\tau$ from $3$ to $6$ improves \textit{DS}, \textit{SR}, and \textit{DE} by $7.00$, $3.04$ percentage points, and $20.16$, respectively. Increasing $\tau$ from $9$ to $12$ instead decreases \textit{DS} and \textit{SR} by $2.20$ and $0.76$ percentage points, indicating increased long-horizon uncertainty. We therefore use $m=5$ and $\tau=6$ as the default settings, balancing driving performance, temporal consistency, and computational cost.

% \subsection{Real-world Deployment}
% We further deploy Crisp-Drive on an AGV platform to verify its sim-to-real transfer capability in three real-world driving scenarios: 1) oncoming obstacle avoidance; 2) mid-road static obstacle avoidance; and 3) emergency response to jaywalking pedestrians.

% As shown in Fig~\ref{figure_Qualitative}, Crisp-Drive successfully executes safe and smooth maneuvers across all scenarios while adapting to distinct driving styles. In Scenarios 1 and 2, the vehicle yields to an oncoming vehicle or static obstacle by either performing a proper lane change or stopping, demonstrating cooperative behavior under different style settings. In Scenario 3, when confronted with a pedestrian suddenly crossing the road, the system successfully initiates braking, exhibiting high responsiveness to hazards. These results demonstrate that Crisp-Drive not only can generalize from simulation to the real world but also retains behavioral flexibility across diverse driving preferences in dynamic environments.

\subsection{Real-World Deployment}
We deploy Crisp-Drive on a low-speed automated guided vehicle (AGV) to evaluate its transferability under controlled real-world conditions. The closed-loop evaluation covers three safety-critical task categories: 1) interactions with oncoming vehicles and jaywalking pedestrians; 2) static-obstacle avoidance; and 3) left- and right-turn maneuvers.

Table~\ref{tab:real_world_evaluation} reports 20 trials for each style--task combination, yielding 180 trials in total. Crisp-Drive achieves success rates of $90.00\%$, $95.00\%$, and $91.67\%$ for the Dynamic, Static, and Turning tasks, respectively, with seven collisions. The Conservative policy provides the largest average minimum clearance ($1.41\,\mathrm{m}$) and the lowest jerk ($0.54$), while the Moderate policy achieves the highest overall success rate ($95.00\%$). The Aggressive policy exhibits more proactive behavior, with a smaller clearance ($0.94\,\mathrm{m}$) and higher jerk ($0.82$), while retaining an overall success rate of $88.33\%$.

Figure~\ref{figure_Qualitative} presents representative executions for oncoming-vehicle interaction, static-obstacle avoidance, pedestrian emergency braking, and turning. The resulting trajectories demonstrate that Crisp-Drive performs scene-appropriate maneuvers while preserving distinguishable Conservative and Aggressive behaviors.

\begin{table}[t]
\centering
\caption{Real-World Evaluation Across Safety-Critical Scenarios and Driving Styles, With $20$ Trials per Style--Scenario Setting.}
\label{tab:real_world_evaluation}
\footnotesize
\setlength{\tabcolsep}{3.2pt}
\renewcommand{\arraystretch}{0.92}
\begin{tabular}{l c c c c c}
\toprule
Scenario & Style & Success$\uparrow$ & Coll.$\downarrow$ & Dist.$\uparrow$ & Jerk$\downarrow$ \\
\midrule
\multirow{3}{*}{Dynamic} & $\mathcal{C}$ & 19 & 1 & $1.42 \pm 0.18$ & $0.61 \pm 0.09$ \\
& $\mathcal{M}$ & 18 & 1 & $1.18 \pm 0.21$ & $0.74 \pm 0.12$ \\
& $\mathcal{A}$ & 17 & 2 & $0.93 \pm 0.16$ & $0.91 \pm 0.15$ \\

\midrule

\multirow{3}{*}{Static} & $\mathcal{C}$ & 19 & 0 & $1.55 \pm 0.16$ & $0.47 \pm 0.07$ \\
& $\mathcal{M}$ & 20 & 0 & $1.28 \pm 0.19$ & $0.58 \pm 0.09$ \\
& $\mathcal{A}$ & 18 & 2 & $1.02 \pm 0.14$ & $0.72 \pm 0.11$ \\

\midrule

\multirow{3}{*}{Turning} & $\mathcal{C}$ & 18 & 0 & $1.26 \pm 0.20$ & $0.55 \pm 0.08$ \\
& $\mathcal{M}$ & 19 & 0 & $1.09 \pm 0.17$ & $0.68 \pm 0.10$ \\
& $\mathcal{A}$ & 18 & 1 & $0.88 \pm 0.15$ & $0.82 \pm 0.13$ \\
\bottomrule
\end{tabular}
\end{table}

\section{Conclusion}
In this work, we present Crisp-Drive, a world model-based reinforcement learning framework for end-to-end driving that addresses three critical limitations of existing approaches: temporal inconsistency in imagined rollouts, suboptimal modeling of agent-environment interactions, and inflexible policy adaptation across diverse driving styles. To this end, Crisp-Drive introduces three key innovations: (i) a temporal consistency regularization that leverages gated cross-attention over historical latent states to stabilize long-horizon dynamics and mitigate error accumulation; (ii) an explicit state disentanglement module that isolates ego-relevant states from background dynamics, thereby enhancing interpretability and risk awareness in decision-making; and (iii) a multi-style policy optimization strategy based on Group Relative Policy Optimization, which enables joint training of diverse driving behaviors through intra-group relative advantage estimation. Evaluated on the Bench2Drive benchmark under closed-loop settings, Crisp-Drive achieves a driving score of \textbf{$84.12$} and a success rate of \textbf{$63.21\%$}, outperforming the previous best world model-based method. Furthermore, we demonstrate successful sim-to-real transfer by deploying Crisp-Drive on a real automated guided vehicle platform, where it exhibits robust performance in dynamic traffic interactions.

\bibliographystyle{IEEEtran}
\bibliography{IEEEabrv,ref1}

\end{document}